\documentclass{article}
\usepackage{float}
\usepackage[english]{babel}
\usepackage{array}
\usepackage[letterpaper,top=2cm,bottom=2cm,left=3cm,right=3cm,marginparwidth=1.75cm]{geometry}
\usepackage{CJKutf8}
\usepackage{amssymb}
\usepackage{booktabs}
\usepackage{authblk}
\usepackage{makecell}

\usepackage{graphicx}

\usepackage{amsmath}
\usepackage{graphicx}
\usepackage[colorlinks=true, allcolors=blue]{hyperref}

\title{Learning latent progression states from spatial heterogeneity in uterine histopathology}
\author[1,5,+,*]{Qiming He}
\author[3,+]{Yan Liu}
\author[2,4,+]{Shuang Ge}
\author[2]{Fan Yang}
\author[3]{Yuxiang Wang}
\author[2]{Ieng Man Zhang}
\author[3]{Jing Yang}
\author[1]{Zihao Jia}
\author[3]{Ajin Hu}
\author[7]{Yexing Zhang}
\author[3]{Zixiu Song}
\author[2]{Qiang Huang}
\author[3]{Xiaoya Zhao}
\author[2]{Zihan Wang}
\author[3]{Xianjing Zheng}
\author[3]{Yijun Zheng}
\author[3]{Liling Lin}
\author[3]{Shuxing Liu}
\author[3]{Bin Bao}
\author[3]{Yue Xie}
\author[2,*]{Tian Guan}
\author[2,5,6*]{Yonghong He}
\author[3,*]{Congrong Liu}

\affil[1]{Fuzhou University Affiliated Provincial Hospital, Interdisciplinary Institute for Medical Engineering, Fuzhou University, Fuzhou, China}
\affil[2]{Institute of Biopharmaceutical and Health Engineering, Tsinghua Shenzhen International Graduate School, Tsinghua University, Shenzhen, China}
\affil[3]{Department of Pathology, Third Hospital, School of Basic Medical Sciences, Peking University Health Science Center, Beijing, China}
\affil[4]{Peng Cheng Laboratory, Shenzhen, China}
\affil[5]{Medical Optical Technology R\&D Center, Research Institute of Tsinghua, Pearl River Delta, Guangzhou, China}
\affil[6]{Jinfeng Laboratory, Chongqing, China}
\affil[7]{Jinan Inspur Data Technology Co., Ltd., Jinan, China}
\affil[+]{Qiming He, Yan Liu, and Shuang Ge contribute equally to this work.}
\affil[*]{Corresponding authors: he-qm@fzu.edu.cn; guantian@sz.tsinghua.edu.cn; Congrong\_liu@163.com; heyh@sz.tsinghua.edu.cn}

\begin{document}
\begin{CJK*}{UTF8}{gbsn}
\date{}
\maketitle

\begin{abstract}
Tumor progression is accompanied by changes in architecture, morphology and microenvironmental organization, yet progression-associated heterogeneity is usually compressed into static diagnostic categories in histopathology. Here we present \textbf{SpaTIE}, a uterus-specific computational pathology framework that learns morphology-aware representations and organizes spatial histopathological heterogeneity into progression-associated tumor states. SpaTIE was developed using 10,426 uterine hematoxylin and eosin whole-slide images and evaluated in TCGA-UCEC and TCGA-UCS cohorts. The learned representations formed morphology manifolds, supported diagnostic, molecular and survival-related prediction tasks, and localized attention to informative tumor regions. Beyond supervised prediction, SpaTIE inferred tumor-state axes from cross-sectional morphology without temporal or molecular supervision. These morphology-derived states were spatially coherent and showed associations with clinicopathological variables and survival outcomes, while not simply recapitulating staging or diagnostic labels. Integrative multi-omics analyses linked the inferred states to DNA methylation, somatic copy-number variation, mutation, RNA-seq and RPPA profiles, highlighting molecular programs related to chromatin regulation, copy-number-associated structural variation, receptor tyrosine kinase signaling, cell adhesion, extracellular-matrix remodeling and metabolic adaptation. Progression-guided virtual perturbation further prioritized molecular features coupled to the morphology-derived state organization. Together, these findings suggest that uterine histopathology contains recoverable progression-associated tumor-state information and establish SpaTIE as a framework for connecting spatial morphology with multi-omics-informed tumor-state discovery.

\end{abstract}
\vspace{0.5em}
\vspace{1em}

\newpage
\section{Introduction}

Tumor progression is not a discrete event but a continuum of coordinated changes in epithelial architecture, cellular morphology, stromal organization, immune contexture and tumor--microenvironment interactions \cite{hanahan2022hallmarks,greaves2012clonal}. In uterine malignancies, such changes are particularly heterogeneous, spanning indolent endometrioid tumors, high-grade carcinomas and aggressive carcinosarcomas with biphasic epithelial--mesenchymal features \cite{tcga2013integrated,cherniack2017integrated}. Although routine hematoxylin and eosin (H\&E) histopathology remains the cornerstone of diagnosis, grading and clinical decision-making, it is usually interpreted through discrete categories such as histological subtype, tumor grade and stage. These categories are clinically useful, but they compress a continuum of morphological and biological variation into static diagnostic labels. Sequencing approaches can generate multi-omics data to characterize the molecular trajectories underlying tumor progression; however, these techniques are associated with substantially high costs. As a result, progression-associated tumor states and within-category heterogeneity may remain under-recognized.

A central but underexplored question is whether routine histopathology contains latent information about progression-associated tumor organization beyond these conventional labels. Tumors are spatially heterogeneous tissues in which morphologically distinct regions often coexist within the same specimen, including areas with preserved glandular differentiation, solid growth, stromal reaction, necrosis, inflammatory infiltration and invasive or dedifferentiated phenotypes \cite{heindl2015relevance,marusyk2012intra}. Rather than representing only sampling noise, such spatial heterogeneity may reflect coexisting tumor states distributed along a progression-associated morphological axis. This concept parallels trajectory inference in single-cell biology, where asynchronous cellular states sampled at one time point can be computationally ordered to recover latent biological processes \cite{trapnell2014dynamics,saelens2019comparison,street2018slingshot}. Recent studies in spatial transcriptomics and single-cell genomics further suggest that tissue organization may preserve latent developmental or pathological state structure \cite{stahl2016visualization,longo2021integrating}. However, whether analogous progression-associated tumor states can be extracted from routine histopathology images, without longitudinal sampling or molecular supervision, remains largely unresolved.

Recent advances in computational pathology and self-supervised representation learning have enabled large-scale extraction of morphological features from whole-slide images (WSIs) \cite{campanella2019clinical,chen2024uni,xu2024gigapath,vorontsov2024virchow,he2026glopath,zhu2025subspecialty}. Pathology foundation models can learn transferable tissue representations and support diverse downstream tasks, including diagnosis, molecular prediction and prognosis \cite{lu2021data,vorontsov2024foundation,ding2025titan,neidlinger2025benchmarking,campanella2025clinical}. Nevertheless, most existing approaches still treat histopathology as a static prediction problem, optimizing representations for classification or regression rather than resolving continuous disease-state organization. In parallel, trajectory inference methods have been widely used in single-cell and spatial omics to reconstruct dynamic biological processes, but their application to routine histopathology remains limited \cite{bergen2020generalizing,deltadahl2025deep,liu2025image,qiu2025quantifying}. This gap motivates a trajectory-aware computational pathology framework that can transform spatial morphological heterogeneity into clinically and biologically interpretable progression-associated tumor states.

\begin{figure}[H]
    \centering
    \includegraphics[width=1\linewidth]{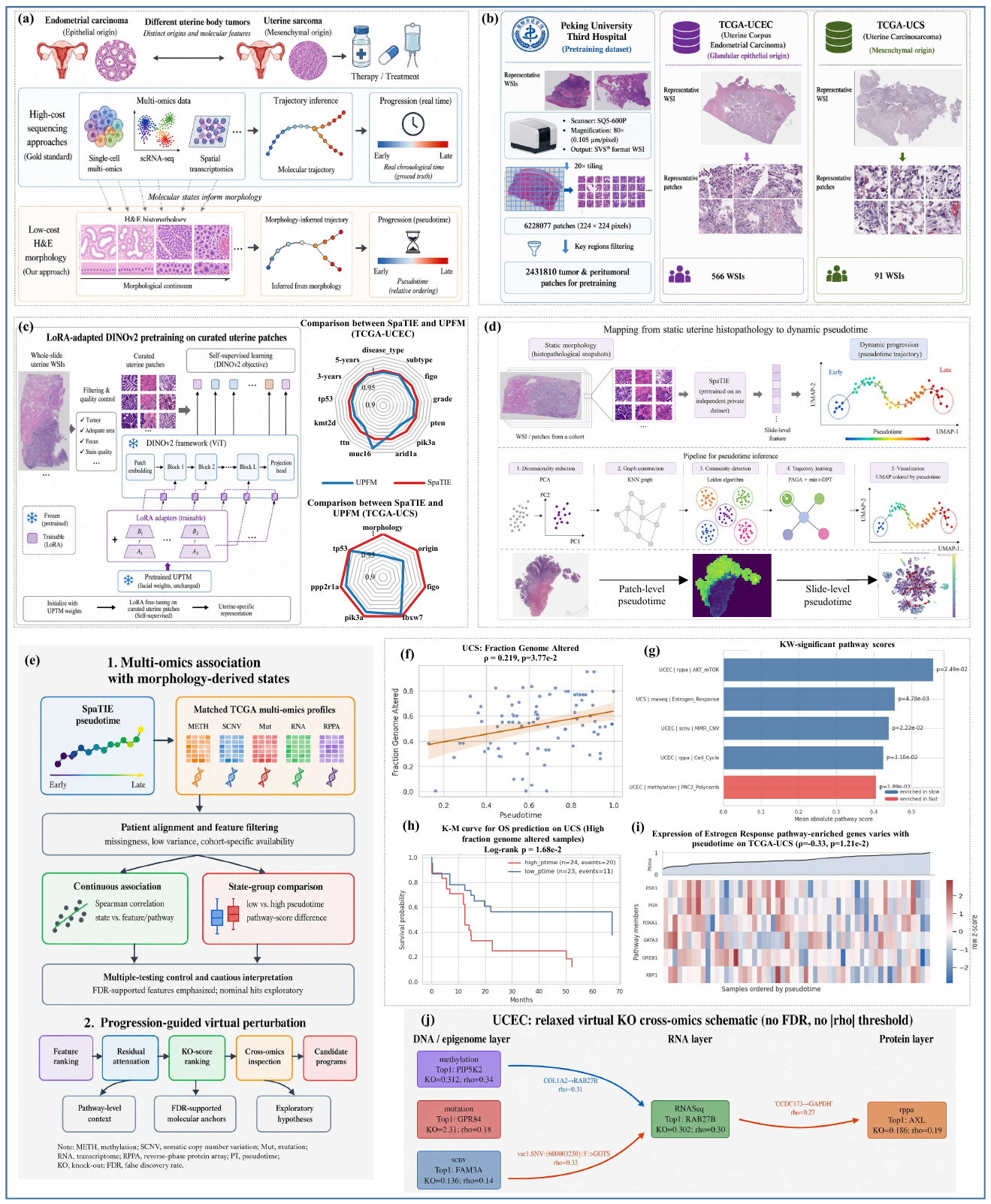}
    \caption{Overview of SpaTIE. \textbf{a, }H\&E histopathology provides spatial morphology that may contain latent information. \textbf{b, }Study cohorts and data resources, including the PK3-Uter pretraining cohort and independent TCGA-UCEC and TCGA-UCS cohorts used for evaluation. \textbf{c, }SpaTIE performed uterus-specific representation learning and beat universal pathology foundation model (UPFM) across downstream clinical tasks. \textbf{d, }Mapping from static histopathology to progression states (pseudotime). \textbf{e, }Analysis workflow for linking SpaTIE-derived progression states to multi-omics features. \textbf{f, }Association between SpaTIE-derived progression states and fraction genome altered in TCGA-UCS. \textbf{g, }Summary of pathway-level associations with low- versus high-pseudotime groups across cohorts and omics modalities. \textbf{h, }Example survival stratification by pseudotime group in the TCGA-UCS high-FGA subgroup. \textbf{i, }Representative state-ordered heatmap of estrogen-response RNA-seq features in TCGA-UCS. \textbf{j, }Cross-omics virtual-perturbation schematic illustrating candidate relationships among epigenomic, transcriptomic and protein-level features.}
    \label{overview}
\end{figure}

Here, we present \textbf{SpaTIE}, designed to identify latent progression-associated states from spatial morphological heterogeneity in uterine histopathology (Fig. \ref{overview}a). We first formulated a theoretical basis showing how coexisting morphological states within histopathology images give rise to progression-associated tumor-state organization (Sec. \ref{conception}). Building on this rationale, SpaTIE learns uterus-specific morphological representations from 10,426 H\&E WSIs and organizes cross-sectional morphology into continuous tumor-state axes without temporal or molecular supervision (Fig. \ref{overview}b-d). We evaluated SpaTIE using large-scale uterine pathology data and independent public cohorts, covering endometrial carcinoma and uterine carcinosarcoma. SpaTIE identified spatially coherent progression-associated organization, and the inferred states showed associations with clinicopathological variables, molecular alterations and survival outcomes (Fig. \ref{overview}f,h). Recent foundation models have demonstrated that histopathological morphology can be aligned with spatial transcriptomic representations at scale \cite{chen2025visual}. Beyond clinicopathological associations, SpaTIE-derived states were linked to multi-omics variation across DNA methylation, somatic copy-number variation, mutation, transcriptomic and proteomic profiles (Fig. \ref{overview}e,g,i). Trajectory-associated molecular analyses and progression-guided virtual perturbation further revealed candidate regulatory programs and pathways related to epithelial--mesenchymal transition, chromatin dysregulation, metabolic adaptation and microenvironmental remodeling (Fig. \ref{overview}j). Together, these findings suggest that routine histopathology contains recoverable progression-associated tumor-state information, providing a framework for trajectory-aware computational pathology and multi-omics-informed characterization of tumor states from H\&E morphology.

\section{Results}

\subsection{Conceptual foundation}\label{conception}
Tumor progression is a continuous but spatially asynchronous process. Within a single histopathology specimen, tumor regions often exhibit substantial morphological diversity, including preserved glandular differentiation, solid growth, stromal activation, necrosis, inflammatory infiltration, and invasive or dedifferentiated phenotypes. Rather than representing only sampling variability, these coexisting patterns may reflect tumor states distributed along a progression-associated morphological continuum \cite{greaves2012clonal,marusyk2012intra}. We therefore reasoned that spatial heterogeneity in routine histopathology may preserve information about disease-state organization that is not fully captured by a single diagnostic label.

This concept is analogous to trajectory inference in single-cell biology, where asynchronously sampled cellular states can be computationally organized to recover latent biological processes \cite{trapnell2014dynamics,saelens2019comparison,street2018slingshot}. We hypothesized that histopathological regions with similar progression-associated morphology should occupy nearby positions in a representation space and, across regions and patients, form a continuous tumor-state organization. Under this view, whole-slide histopathology images are spatial snapshots containing mixtures of coexisting tumor states, rather than single homogeneous disease categories.

To formalize this intuition, let $x \in \mathcal{X}$ denote a local histopathological observation, such as an image patch, and let $t \in \mathcal{T}$ represent an unobserved progression-associated tumor state. We assume that tissue morphology arises from a stochastic generative process:
\begin{equation}
x = g(t,\epsilon), \quad \epsilon \sim p(\epsilon),
\end{equation}
where $g(\cdot)$ describes progression-associated morphological variation and $\epsilon$ captures patient-specific variability, local microenvironmental differences, and imaging noise. Under this formulation, histopathology specimens consist of spatially distributed observations sampled from multiple latent tumor states:
\begin{equation}
p(x)=\int_{\mathcal{T}} p(x|t)p(t)\,dt.
\end{equation}

Recent advances in self-supervised representation learning suggest that morphology-aware encoders can preserve high-level tissue organization within latent embedding spaces \cite{chen2024uni,xu2024gigapath,vorontsov2024foundation}. Let $z_i$ denote the latent representation of observation $x_i$. If progression-associated morphological similarity is preserved after representation learning, observations arising from nearby tumor states should remain locally close in the embedding space. This local organization can be characterized through a similarity kernel:
\begin{equation}
K_{ij}=\exp\left(-\frac{\|z_i-z_j\|^2}{\sigma^2}\right),
\end{equation}
which defines a neighborhood structure over tissue regions. Diffusion- or graph-based trajectory inference methods can then be applied to organize tissue regions along low-dimensional state continua \cite{coifman2005geometric,haghverdi2016diffusion}. Under this framework, latent progression-associated states emerge as geometric organization within morphology-aware representation spaces, rather than as explicit temporal measurements.

Motivated by this biological and computational rationale, we developed SpaTIE, a trajectory-aware computational pathology framework designed to recover progression-associated tumor states from uterine histopathology. SpaTIE combines uterus-specific representation learning with tumor-state inference to characterize morphology-associated organization across tissue regions and patients without requiring temporal sampling or molecular supervision.

\subsection{Overview of SpaTIE}

An overview of SpaTIE is illustrated in Fig. \ref{overview}. Tumor progression is commonly studied using molecular profiling approaches, which can provide biologically informative state trajectories but remain difficult to apply at scale in routine pathology (Fig. \ref{overview}a). In contrast, H\&E histopathology is widely available and inherently reflects spatial morphological heterogeneity within tumor tissues. SpaTIE was designed based on the hypothesis that these spatially coexisting morphological states may preserve progression-associated organization that can be recovered computationally.
To support morphology-aware state modeling, we first constructed a large-scale uterine histopathology resource (Fig. \ref{overview}b). Specifically, we collected 10,426 WSIs from Peking University Third Hospital, Beijing, China for uterus-specific self-supervised pretraining (PK3-Uter), covering diverse uterine pathological subtypes and morphological patterns. In addition, TCGA-UCEC and TCGA-UCS cohorts were incorporated as independent external datasets for clinicopathological evaluation, survival analysis, and multi-omics association studies \cite{tcga2013integrated}.

SpaTIE performs uterus-specific morphological representation learning using a pathology foundation model adapted through DINOv2-style self-supervised learning and LoRA-based efficient fine-tuning \cite{oquab2023dinov2,hu2022lora} (Fig. \ref{overview}c). Through this process, histopathological regions with similar morphology and progression-associated characteristics become locally organized within a shared latent embedding space. In feature-space comparisons, SpaTIE showed more efficient manifold construction for uterine disease types than a universal pathology foundation model (UPFM) pretrained on multi-organ and multi-disease datasets \cite{chen2024towards}. On TCGA-UCEC and TCGA-UCS, SpaTIE also improved or matched UPFM across downstream tasks covering morphological diagnosis, gene prediction, and survival prediction (Sec. \ref{detail_static}).

Building upon the learned embedding space, SpaTIE organizes cross-sectional histopathology into latent progression-associated tumor-state axes without requiring temporal sampling or molecular supervision (Fig. \ref{overview}d, Sec. \ref{detail_progression}). Morphology-aware embeddings are arranged into continuous manifolds, enabling slide-level and region-level characterization of tumor-state organization across tissues and patients. The inferred states are then linked to clinicopathological variables, molecular alterations, and spatial morphology patterns.
To evaluate the biological and clinical relevance of morphology-derived tumor states, we analyzed their associations with histological subtype, tumor grade, genomic alteration burden, prognosis, and patient survival (Fig. \ref{overview}f,h; Sec. \ref{detail_progression}). Beyond clinicopathological analyses, we further integrated these states with multi-omics profiles, including DNA methylation, somatic copy-number variation, mutation, transcriptomic, and proteomic data (Fig. \ref{overview}e,g,i; Sec. \ref{detail_simple}). Trajectory-aware molecular analyses and progression-guided virtual perturbation further identified candidate pathways and regulatory programs potentially associated with morphology-derived tumor-state organization in uterine tumors (Fig. \ref{overview}j; Sec. \ref{detail_ko}).

\subsection{SpaTIE learns a static morphology-aware representation space}\label{detail_static}

\begin{figure}[H]
    \centering
    \includegraphics[width=1\linewidth]{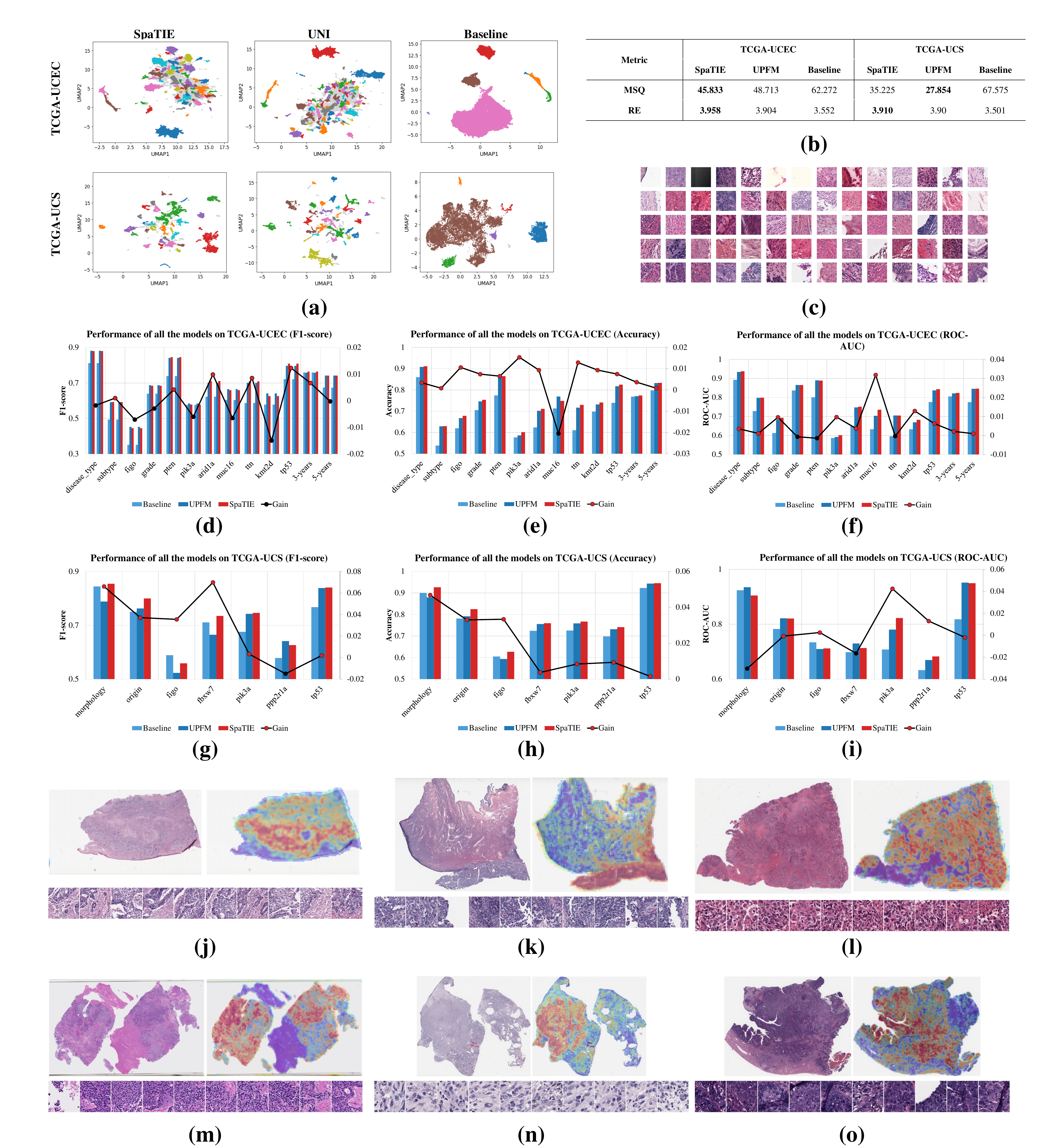}
    \caption{SpaTIE learns a static morphology-aware representation space for uterine histopathology. \textbf{a, }UMAP visualization of patch-level embeddings extracted by SpaTIE, UPFM, and Baseline on TCGA-UCEC and TCGA-UCS. \textbf{b, }Quantitative comparison of static latent-space organization. \textbf{c, }Examples of representative cluster-center patches identified from the SpaTIE latent space (TCGA-UCEC). \textbf{d-f, }Performance comparison of downstream prediction tasks on TCGA-UCEC using F1-score, accuracy, and ROC-AUC. \textbf{g-i, }Performance comparison of downstream prediction tasks on TCGA-UCS using F1-score, accuracy, and ROC-AUC. \textbf{j-l, }Representative attention maps and high-attention histopathological regions from SpaTIE-based downstream prediction models for disease\_type, pten, and tp53 respectively (TCGA-UCEC). \textbf{m-o, }Representative attention maps and high-attention histopathological regions from SpaTIE-based downstream prediction models for morphology, tp53, and origin respectively (TCGA-UCS).}
    \label{static}
\end{figure}

Before inferring progression-associated states, we first examined whether SpaTIE could construct a biologically meaningful static representation space for uterine histopathology. We compared SpaTIE with UPFM and baseline representations on TCGA-UCEC and TCGA-UCS. UMAP visualization of patch-level embeddings showed that SpaTIE organized uterine histopathological patches into more structured and fine-grained latent manifolds than the baseline model, while maintaining local separation of morphologically distinct tissue regions (Fig. \ref{static}a). Compared with UPFM, SpaTIE preserved a similarly rich but more uterus-adapted embedding structure, suggesting that domain-adapted self-supervised learning improved the representation of uterine-specific morphology.

We next quantified the quality of the static representation space using manifold structure quality (MSQ) and representation entropy (RE) (see details in Sec. \ref{metric_static}). SpaTIE achieved the best overall performance across the two cohorts, with MSQ values of 45.833 and 35.225 and RE values of 3.958 and 3.910 on TCGA-UCEC and TCGA-UCS, respectively (Fig. \ref{static}b). These results indicate that SpaTIE improves both structural organization and information distribution in the latent histopathological representation, providing a stable feature basis for subsequent tumor-state inference.

To further interpret the learned representation space, we inspected representative cluster-center patches derived from the SpaTIE embeddings (Fig. \ref{static}c). These cluster centers covered diverse tissue and tumor morphologies, including tumor epithelium, stromal regions, necrotic areas, inflammatory components, tissue edges, and morphologically heterogeneous tumor regions. SpaTIE not only separated broad tissue compartments but also captured finer variation within tumor regions, such as glandular architecture, solid growth, and poorly differentiated patterns. This suggests that the static representation space learned by SpaTIE reflects biologically relevant morphological organization rather than only low-level visual similarity.

We then evaluated whether SpaTIE-derived static features retained clinically useful information through downstream prediction tasks (see details in Supplementary Sec. \ref{sm_tasks} and Supplementary Sec. \ref{sm_downstream}). On TCGA-UCEC, SpaTIE achieved competitive performance across multiple tasks, including histological grade prediction, subtype classification, molecular alteration prediction, and survival-related prediction, as measured by F1-score, accuracy, and ROC-AUC (Fig. \ref{static}d-f). SpaTIE showed strong performance for grade-related tasks and mutation prediction of key endometrial cancer genes such as PTEN, TP53, and ARID1A, whereas tasks such as FIGO stage prediction remained more challenging. On TCGA-UCS, SpaTIE also achieved robust performance in morphology-related and molecular prediction tasks, including morphology classification, origin prediction, and TP53 mutation prediction (Fig. \ref{static}g-i). These results indicate that SpaTIE-derived static morphological features encode clinicopathological and molecular information relevant to uterine malignancies.

Finally, we visualized attention maps from SpaTIE-based downstream prediction models to assess whether model decisions were supported by histopathologically meaningful regions. Representative heatmaps showed that high-attention regions were concentrated in tumor-rich or morphologically informative areas rather than background tissue or irrelevant regions (Fig. \ref{static}j--o). High-attention patches further corresponded to regions with diagnostic or prognostic morphology, supporting the interpretability of SpaTIE-derived static features. Together, these analyses show that SpaTIE constructs a static morphology-aware representation space that is structurally coherent, biologically interpretable, and clinically informative.

\subsection{SpaTIE maps static morphology to latent progression-associated tumor states}\label{detail_progression}

\begin{figure}[H]
    \centering
    \includegraphics[width=0.8\linewidth]{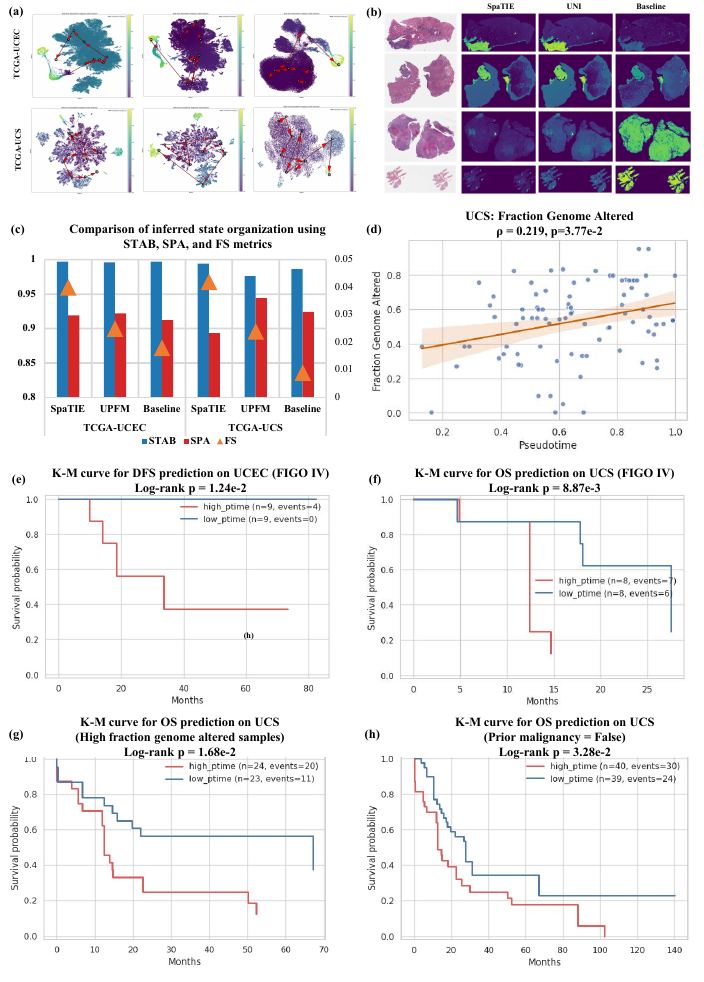}
    \caption{
SpaTIE maps static morphological representations to latent progression-associated tumor states.
(a) UMAP visualization of latent state organization inferred from SpaTIE, UPFM, and baseline embeddings on TCGA-UCEC and TCGA-UCS.
(b) Representative spatial tumor-state maps generated by SpaTIE, UPFM, and baseline models.
(c) Quantitative comparison of inferred state organization using STAB, SPA, and FS metrics.
(d) Association between SpaTIE-derived states and fraction genome altered in TCGA-UCS.
(e,f) Exploratory survival stratification within FIGO stage IV patients in TCGA-UCEC and TCGA-UCS.
(g,h) Exploratory subgroup survival analyses in TCGA-UCS patients with high fraction genome altered or without prior malignancy.
}
    \label{ptime}
\end{figure}

Having established a uterus-adapted morphology space, we next asked whether static histopathological features could be organized into latent progression-associated tumor states. We applied trajectory-aware inference to SpaTIE-derived embeddings and obtained continuous state axes without temporal labels or molecular supervision. UMAP visualization showed coherent latent state structures in both TCGA-UCEC and TCGA-UCS (Fig. \ref{ptime}a). We further examined the spatial distribution of latent states on representative WSIs. SpaTIE generated spatial state maps that were more locally coherent and morphologically consistent than those produced by UPFM and baseline models (Fig. \ref{ptime}b). Regions with similar histological appearance tended to share similar state values, whereas morphologically distinct regions showed different state levels. In contrast, the baseline model produced less organized spatial maps and frequently failed to distinguish tumor-rich regions from non-tumor or stromal regions. These spatial results suggest that the inferred states are linked to tissue-level morphology, rather than being only artifacts of dimensionality reduction.

We evaluated the inferred states using stability (STAB), spatial consistency (SPA), and feature smoothness (FS) (see details in Sec. \ref{metric_ptime}). SpaTIE achieved consistently strong performance across these metrics, with STAB values of 0.997 and 0.994, SPA values of 0.919 and 0.893, and FS values of 0.040 and 0.042 on TCGA-UCEC and TCGA-UCS, respectively (Fig. \ref{ptime}c). These results indicate that the SpaTIE-derived axes are stable and locally coherent, supporting the presence of progression-associated organization within the learned morphology manifold.

To assess the clinicopathological relevance of SpaTIE-derived states, we analyzed slide-level state values against available clinical annotations. In TCGA-UCEC, inferred states were associated with FIGO stage (Kruskal--Wallis $H = 15.357$, $p = 1.54 \times 10^{-3}$, FDR = 0.0077), fraction genome altered (FGA; $\rho = -0.181$, $p = 1.76 \times 10^{-5}$, FDR = $5.29 \times 10^{-5}$), and prior malignancy status ($p = 0.0078$, FDR = 0.0259). Extended GDC clinical annotations further showed associations with tumor grade ($H = 35.142$, $p = 1.14 \times 10^{-7}$, FDR = $5.68 \times 10^{-6}$), percent tumor invasion ($\rho = -0.225$, $p = 3.98 \times 10^{-7}$, FDR = $9.96 \times 10^{-6}$), and disease response ($H = 12.829$, $p = 0.0016$, FDR = 0.0238). By contrast, disease type, primary diagnosis, and morphology code were not significant after multiple-testing correction in TCGA-UCEC. In TCGA-UCS, the inferred state axis showed a positive association with FGA ($\rho = 0.219$, $p = 3.77 \times 10^{-2}$, FDR = 0.113; Fig. \ref{ptime}d), consistent with a link between morphology-derived states and genomic alteration burden. By contrast, FIGO stage and ICD-10 anatomical classification were not significant in TCGA-UCS (FIGO: $p = 0.424$, FDR = 0.566; ICD-10: $p = 0.292$, FDR = 0.467). These results indicate that the inferred states capture selected tumor-burden and morphology-associated signals, rather than broadly tracking every available clinical annotation (see details in Table \ref{sm_ptime_clinic}).

We next evaluated prognostic associations. In TCGA-UCEC, slide-level survival analysis showed that the state axis was associated with overall survival (log-rank $p = 0.0038$; Cox HR = 0.677, 95\% CI 0.536--0.855, $p = 0.0010$). Disease-free survival was less consistently associated in the full UCEC cohort, but an FIGO stage IV analysis showed separation of disease-free survival by state group ($n = 18$, log-rank $p = 0.0124$; Fig. \ref{ptime}e). In TCGA-UCS, several subgroup analyses showed prognostic separation. Within FIGO stage IV samples, state groups differed in overall survival ($n = 16$, log-rank $p = 8.87 \times 10^{-3}$; Fig. \ref{ptime}f). Among UCS patients with high FGA, the state axis also stratified overall survival ($n = 47$, log-rank $p = 1.68 \times 10^{-2}$ in the analysis shown in Fig. \ref{ptime}g), consistent with the observed state--FGA correlation. A further nominal association was observed among UCS patients without prior malignancy ($n = 79$, log-rank $p = 3.28 \times 10^{-2}$; Fig. \ref{ptime}h). These subgroup signals should be interpreted as exploratory, particularly in UCS where sample sizes are modest. Taken together, SpaTIE maps uterine histopathology into spatially coherent latent states that are linked to selected clinicopathological variables, genomic alteration burden, and prognostic patterns in UCEC and UCS, without implying a universal or deterministic timeline of clinical progression.

\subsection{Progression-associated states are linked to pathway activity and multi-omics features}\label{detail_simple}

\begin{figure}[H]
    \centering
    \includegraphics[width=1\linewidth]{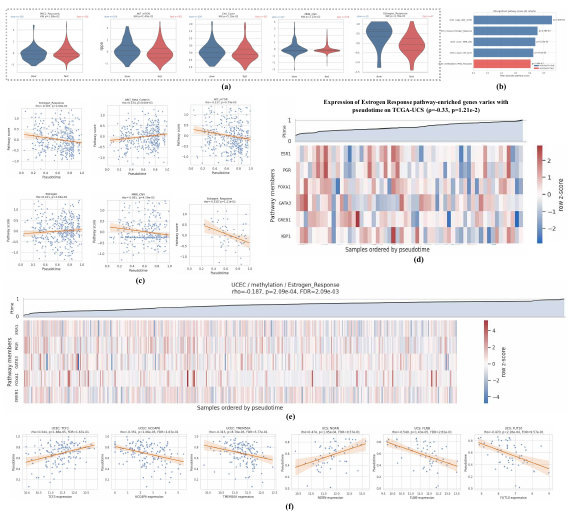}
    \caption{
Association between SpaTIE-derived progression-associated states and pathway or molecular features.
(a) Violin plots comparing pathway scores between low- and high-state groups for significant pathway-level associations.
(b) Summary of significant pathway associations across cohorts and omics modalities.
(c) Representative scatter plots showing correlations between inferred states and pathway scores.
(d) State-ordered heatmap of UCS RNA-seq estrogen-response pathway members.
(e) State-ordered heatmap of UCEC methylation features in the estrogen-response pathway.
(f) Representative feature-level associations between inferred states and individual molecular features.
}
    \label{simple_association}
\end{figure}

We next asked whether morphology-derived states could be molecularly contextualized. SpaTIE-derived state values were integrated with pathway scores and feature-level multi-omics profiles across TCGA-UCEC and TCGA-UCS. For pathway-level analysis, patients were divided into low- and high-state groups according to the inferred state values, and pathway-score distributions were compared across DNA methylation, somatic copy-number variation (SCNV), mutation, RNA-seq, and RPPA modalities. A limited set of pathway programs differed between state groups (Fig. \ref{simple_association}a,b). In TCGA-UCEC, associations were observed for PRC2/Polycomb-related methylation ($p = 0.019$), MMR-related SCNV ($p = 0.022$), AKT--mTOR RPPA signaling ($p = 0.025$), and cell-cycle-related RPPA signaling ($p = 0.031$). In TCGA-UCS, RNA-seq-based estrogen-response pathway scores differed between state groups ($p = 0.0048$). These results suggest that SpaTIE-derived states are linked to selected molecular programs rather than isolated image-derived variation.

We further examined continuous associations between inferred states and pathway scores. Representative examples showed correlations with multiple pathway-level activities, including UCEC methylation-based estrogen-response scores ($\rho = -0.187$, $p = 2.09 \times 10^{-4}$), UCEC methylation-based WNT/$\beta$-catenin scores ($\rho = 0.134$, $p = 8.06 \times 10^{-3}$), UCEC RPPA AKT--mTOR scores ($\rho = -0.137$, $p = 7.74 \times 10^{-3}$), UCEC SCNV MMR scores ($\rho = -0.091$, $p = 4.29 \times 10^{-2}$), and UCS RNA-seq estrogen-response scores ($\rho = -0.330$, $p = 1.21 \times 10^{-2}$; Fig. \ref{simple_association}c). State-ordered heatmaps further showed gradual pathway-member variation along the inferred axis, including estrogen-response-related RNA-seq features in UCS and estrogen-response-related methylation features in UCEC (Fig. \ref{simple_association}d,e). These observations provide pathway-level molecular context for the morphology-derived state organization.

At the feature level, SpaTIE-derived states were associated with selected molecular features across omics modalities (Fig. \ref{simple_association}f and Supplementary Sec. \ref{sm_simple_association}). In TCGA-UCEC, the clearest corrected support was concentrated in methylation, SCNV, and RPPA layers; representative features overlapping with the virtual perturbation analysis included \textit{PRR22}, \textit{RBM27}, and \textit{LRRC42} for methylation, \textit{FAM81A}, \textit{RNA5SP396}, and \textit{GCNT3} for SCNV, and AXL, GAPDH, ERRFI1, CTNNA1, and SERPINE1 for RPPA. RNA-seq and mutation features in UCEC were interpreted more cautiously because top-ranked candidates had limited FDR support in the current analysis.

TCGA-UCS showed stronger nominal effect sizes for several individual features but weaker FDR support, consistent with the smaller cohort size. Representative UCS RNA-seq associations included \textit{UFD1L}, \textit{FGF1}, \textit{CCNG1}, \textit{IL33}, and \textit{FLNB}, while UCS methylation, mutation, SCNV, and RPPA analyses highlighted additional candidate features with state-associated trends. Collectively, these pathway- and feature-level analyses indicate that morphology-derived states are accompanied by molecular variation across epigenetic, genomic, transcriptomic, and proteomic layers. We interpret these associations as molecular context for the inferred state axis, not as definitive evidence of single-feature molecular drivers.

\subsection{Progression-guided virtual perturbation prioritizes candidate state-associated programs}\label{detail_ko}

To further evaluate whether molecular features were coupled to SpaTIE-derived state organization, we performed progression-guided virtual perturbation analysis across five omics modalities. For each molecular feature, we estimated a virtual perturbation score by measuring how strongly the association between the feature and state value was attenuated after removing the feature-associated component. This analysis was designed as a computational feature-prioritization strategy rather than experimental gene knockout or causal validation, following the emerging paradigm of AI-based virtual cells research, which emphasize in silico perturbation for biological hypothesis generation rather than direct causal inference \cite{bunne2024build}.

\begin{figure}[H]
    \centering
    \includegraphics[width=1\linewidth]{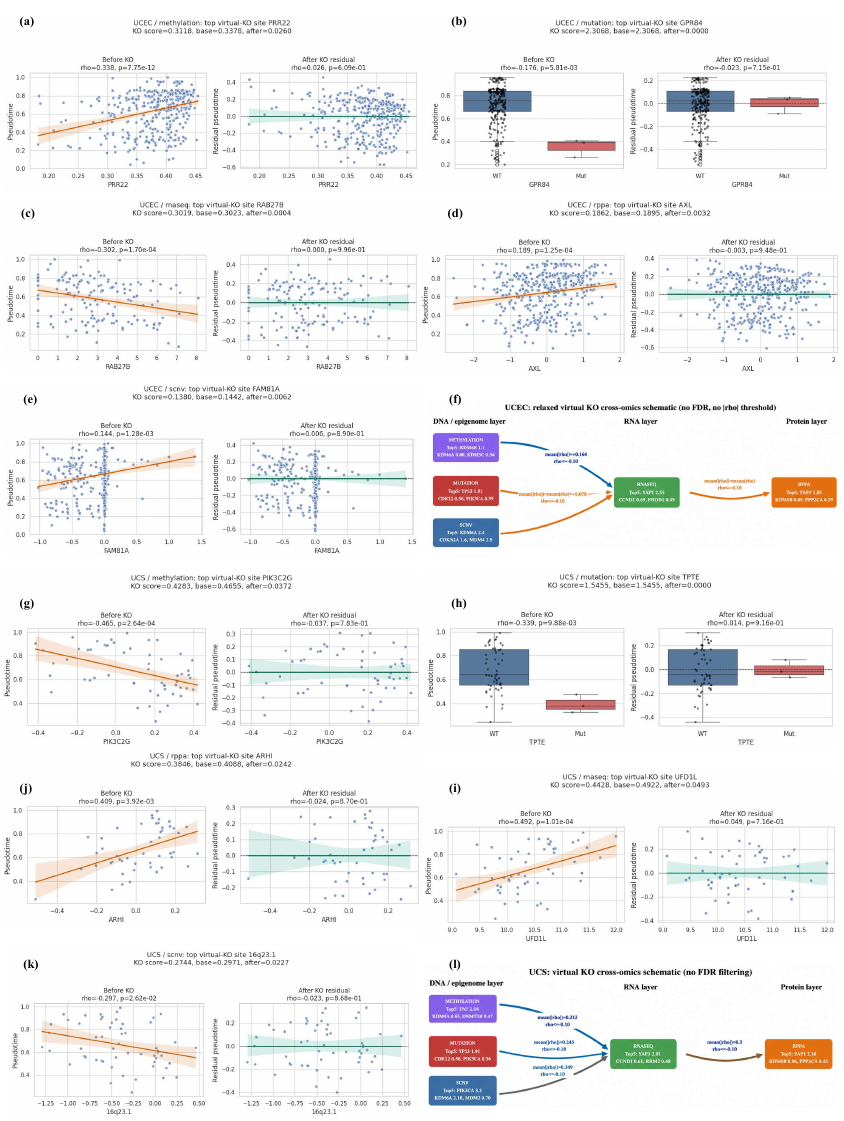}
    \caption{
Progression-guided virtual perturbation analysis.
(a-e) Representative virtual perturbation results in TCGA-UCEC across methylation, mutation, RNA-seq, RPPA, and SCNV modalities.
(f) Cross-omics schematic of UCEC candidate perturbation relationships.
(g-k) Representative virtual perturbation results in TCGA-UCS across methylation, mutation, RNA-seq, RPPA, and SCNV modalities.
(l) Cross-omics schematic of UCS candidate perturbation relationships.
}
    \label{ko}
\end{figure}

In TCGA-UCEC, virtual perturbation highlighted candidate features across epigenetic, genomic, transcriptomic, and proteomic layers (Fig. \ref{ko}a-e). The top methylation-associated perturbation feature was \textit{PRR22} (KO score = 0.312), whose association with state values was markedly reduced after virtual perturbation ($\rho$ from 0.338 to 0.026). Mutation-level analysis prioritized \textit{GPR84} (KO score = 2.307), while RNA-seq analysis identified \textit{RAB27B} as the strongest transcriptomic candidate (KO score = 0.302; $\rho$ from -0.302 to 0.0004). At the protein level, AXL showed the largest RPPA perturbation effect (KO score = 0.186; $\rho$ from 0.189 to 0.003), suggesting a potential link between morphology-derived state organization and receptor tyrosine kinase or EMT-associated signaling. SCNV analysis prioritized \textit{FAM81A} (KO score = 0.138; $\rho$ from 0.144 to 0.006). Across top-ranked UCEC features, FDR-supported state associations were most evident in methylation, SCNV, and RPPA modalities, supporting these layers as relatively robust molecular anchors for the inferred state axis.

In TCGA-UCS, virtual perturbation identified a distinct set of candidate features (Fig. \ref{ko}g-k). The strongest methylation-associated candidate was \textit{PIK3C2G} (KO score = 0.428), whose state association was reduced from $\rho = -0.465$ to $\rho = 0.037$ after perturbation. Mutation-level analysis prioritized \textit{TPTE} (KO score = 1.546), RNA-seq analysis prioritized \textit{UFD1L} (KO score = 0.443; $\rho$ from 0.492 to 0.049), RPPA analysis prioritized ARHI (KO score = 0.385; $\rho$ from 0.409 to 0.024), and SCNV analysis prioritized chromosome region 16q23.1 (KO score = 0.274; $\rho$ from -0.297 to 0.023). Although UCS perturbation scores were often large, individual feature-level FDR support was limited, likely reflecting the smaller sample size of the TCGA-UCS cohort. These UCS candidates were therefore interpreted as exploratory state-associated molecular hypotheses.

Cross-omics integration further suggested that progression-associated state organization reflects distributed molecular coupling rather than a single linear driver (Fig. \ref{ko}f,l). In UCEC, top perturbation candidates did not form a strong same-gene DNA--RNA--protein chain, but instead suggested convergence from methylation and SCNV alterations toward transcriptomic and proteomic programs involving adhesion, signaling, metabolism, and extracellular remodeling. In UCS, several cross-layer relationships connected methylation or mutation candidates to RNA-level features, including same-gene methylation--RNA links such as \textit{MGC87042} and \textit{ROR2}, as well as additional candidate links involving \textit{PRRX2}, \textit{C11orf54}, and \textit{OMA1}. These patterns are compatible with the pronounced genomic and morphological heterogeneity of uterine carcinosarcoma.

Together, progression-guided virtual perturbation supports the utility of SpaTIE-derived states for molecular hypothesis generation (Supplementary Sec. \ref{sm_KO}). The top-ranked candidates point to possible biological themes including epigenetic regulation, copy-number-associated structural variation, receptor tyrosine kinase signaling, cell adhesion, extracellular-matrix remodeling, metabolic adaptation, and stress-response programs. Because the perturbation procedure is computational and based on bulk-level associations, these results should not be interpreted as direct causal evidence. Rather, they prioritize candidate molecular programs that may help explain how morphology-derived state organization is coupled to multi-omics tumor-state variation.

\section{Discussion}

In this study, we developed SpaTIE, a uterus-specific computational pathology framework that connects morphology-aware representation learning with progression-associated tumor-state inference. Rather than using histopathology only as an input for static diagnostic or molecular prediction, SpaTIE organizes spatially heterogeneous H\&E morphology into latent state axes and then relates these axes to clinicopathological variables, survival outcomes, multi-omics profiles and computational perturbation analyses. Across TCGA-UCEC and TCGA-UCS, SpaTIE produced coherent morphology manifolds, supported a range of downstream prediction tasks, localized attention to histopathologically informative regions and revealed tumor-state organization with selected clinical and molecular associations. These findings suggest that uterine histopathology contains recoverable information about tumor-state heterogeneity that extends beyond conventional categorical labels.

A first contribution of SpaTIE is the construction of a uterus-adapted pathology representation. Recent pathology foundation models have demonstrated broad transferability across organs and tasks \cite{chen2024uni,xu2024gigapath,vorontsov2024foundation,ding2025titan}. However, uterine malignancies have distinctive morphological and biological spectra, ranging from low-grade endometrioid carcinoma to high-grade carcinoma and carcinosarcoma with epithelial--mesenchymal features \cite{tcga2013integrated,cherniack2017integrated,crosbie2022endometrial}. By adapting representation learning to a large uterine histopathology resource, SpaTIE captured organ-relevant morphology and maintained competitive performance in diagnostic, molecular and survival-related prediction tasks. Importantly, the value of this adaptation was not limited to conventional task metrics; it also provided a more coherent substrate for downstream state inference.

The second contribution is conceptual. Most computational pathology pipelines treat each WSI as a static object to be assigned a diagnostic, prognostic or molecular label. SpaTIE instead treats tissue morphology as a spatial mixture of coexisting tumor states. This perspective is inspired by trajectory inference in single-cell biology, where asynchronously sampled cells are ordered to recover latent biological processes \cite{trapnell2014dynamics,saelens2019comparison,bergen2020generalizing}. In histopathology, the inferred axes should not be interpreted as direct chronological timelines. Rather, they represent morphology-derived, progression-associated tumor-state organization embedded in cross-sectional tissue images. This distinction is important: the framework seeks to recover structured disease-state variation from spatial morphology, not to prove the temporal history of an individual tumor.

The clinicopathological analyses support this restrained interpretation. SpaTIE-derived states were associated with selected variables related to tumor burden and disease state, including FIGO stage, FGA, tumor grade, tumor invasion, residual disease and disease response. The UCS association between the inferred state axis and FGA is particularly relevant because it links morphology-derived organization to genomic alteration burden. UCS and high-risk subgroup findings suggested potential prognostic relevance in settings such as FIGO stage IV disease and high-FGA tumors. At the same time, the associations were not universal across all clinical variables or endpoints. SpaTIE-derived states should therefore be viewed as partially aligned with clinical progression and outcome, rather than as substitutes for established staging, grading or prognostic models.

A third contribution is the integration of morphology-derived tumor states with multi-omics data. Across DNA methylation, SCNV, mutation, RNA-seq and RPPA profiles, the inferred states were linked to selected molecular programs, including chromatin regulation, copy-number-associated structural variation, receptor tyrosine kinase signaling, cell adhesion, extracellular-matrix remodeling and metabolic adaptation. These analyses are best interpreted as molecular context for the image-derived state axes. The strongest feature-level support was observed in UCEC methylation, SCNV and RPPA analyses, whereas transcriptomic and mutation-level findings, particularly in UCS, were more exploratory. Nevertheless, the convergence of signals across multiple molecular layers suggests that morphology-derived state organization may reflect systems-level tumor biology rather than isolated visual variation.

Progression-guided virtual perturbation further extends this framework toward hypothesis generation. By estimating how strongly feature--state associations were attenuated after computational perturbation, SpaTIE prioritized candidate molecular features coupled to the inferred morphology-derived state axes. This analysis is not equivalent to experimental gene knockout and should not be interpreted as causal validation. Its value is instead practical and exploratory: it ranks candidate features and cross-omics relationships that may help explain why particular molecular programs co-vary with morphology-derived states. In this sense, virtual perturbation provides a bridge from descriptive association analysis to experimentally testable hypotheses.

Several limitations should be acknowledged. First, SpaTIE infers state organization from cross-sectional H\&E cohorts rather than longitudinally sampled tumors. The inferred axes therefore cannot establish chronological tumor evolution, clonal ancestry or branching progression routes. Future studies incorporating longitudinal sampling, multi-region sequencing and spatial molecular profiling will be needed to determine how these morphology-derived states relate to true evolutionary trajectories \cite{marusyk2012intra,longo2021integrating}. Second, although SpaTIE was trained on a large uterine histopathology resource and evaluated in TCGA cohorts, broader external validation is needed across institutions, scanners, staining protocols and patient populations. This is especially important for TCGA-UCS, where sample size is modest and several survival or multi-omics results should be considered exploratory. Third, the molecular analyses are based largely on bulk omics profiles. Bulk methylation, copy-number, transcriptomic and proteomic measurements can provide useful biological context, but they cannot precisely assign molecular programs to the same spatial regions that drive the histopathological state maps. Spatial transcriptomics, multiplex imaging and spatial proteomics could refine the interpretation of these states and help identify the cellular compartments contributing to each morphology-derived axis \cite{stahl2016visualization,shulman2026ai,liu2025spatial}. Fourth, the virtual perturbation analysis remains computational. It prioritizes candidate genes, proteins and genomic features, but it does not prove that these features drive the histological states. Wet-lab validation, such as CRISPR perturbation, organoid or cell-line models, and targeted functional assays, will be required to test whether the nominated candidates mechanistically regulate morphology-associated tumor-state programs.

In summary, SpaTIE provides a uterus-specific framework for connecting predictive pathology with tumor-state discovery. By combining organ-adapted pathology foundation modeling, spatially coherent state inference, multi-omics association analysis and computational perturbation, SpaTIE suggests a route for studying progression-associated tumor heterogeneity from routine histopathology while maintaining a clear boundary between image-derived state organization, molecular association and causal mechanism.


\section{Methods}

\subsection{Ethics approval}

This study was approved by the Medical Ethics Committee of Peking University Third Hospital (Approval No.: IRB00006761-M20250620). Publicly available TCGA-UCEC and TCGA-UCS were obtained from \hyperlink{https://www.cancer.gov/ccg/research/genome-sequencing/tcga}{https://www.cancer.gov/ccg/research/genome-sequencing/tcga}.

\subsection{Study cohorts and data sources}

\subsubsection{PK3-Uter}

To construct a uterus-specific pathology foundation model, we established a large-scale uterine histopathology dataset termed PK3-Uter from Peking University Third Hospital. The cohort included archival uterine histopathology WSIs. The dataset covered multiple uterine pathological entities and morphological patterns, including endometrioid carcinoma, serous carcinoma, clear cell carcinoma, carcinosarcoma, hyperplasia-related lesions, and other uterine pathological subtypes.

All slides were scanned using Shenzhen Shengqiang SQS-600P pathology scanner at 80$\times$ magnification (0.105 $\mu$m/pixel), stored in SDPC format. To facilitate large-scale self-supervised pretraining, tissue regions were first identified using foreground extraction and tissue masking. Tumor-rich regions were subsequently screened and extracted for downstream processing.

For self-supervised pretraining, WSIs were divided into non-overlapping image patches at 20$\times$ magnification. Each patch had a spatial size of 224 $\times$ 224 pixels. In total, the PK3-Uter contained 10,426 WSIs and approximately 62,288,077 image patches after quality control and tissue-region filtering. 

\subsubsection{TCGA-UCEC and TCGA-UCS}

Two independent public uterine malignancy cohorts from The Cancer Genome Atlas were used for external evaluation and multi-omics analysis: uterine corpus endometrial carcinoma (TCGA-UCEC) and uterine carcinosarcoma (TCGA-UCS) \cite{tcga2013integrated,cherniack2017integrated}.

For TCGA-UCEC, we collected H\&E whole-slide images together with clinicopathological annotations, including histological subtype, tumor grade, FIGO stage, ICD anatomical site code, mutation profiles, survival information, and molecular data. Similarly, TCGA-UCS included WSIs and corresponding clinical and molecular annotations for uterine carcinosarcoma samples. WSIs with severe artifacts, incomplete tissue regions, or insufficient tumor content were excluded from downstream analysis. After quality control, the final TCGA-UCEC cohort included 566 WSIs, whereas TCGA-UCS included 91 WSIs. For downstream prediction tasks, patient-level train-validation-test splitting was performed to avoid data leakage between slides originating from the same patient.

\subsubsection{Multi-omics data acquisition and preprocessing}

To investigate the molecular correlates of SpaTIE-derived progression-associated tumor states, we collected multiple molecular modalities from TCGA-UCEC and TCGA-UCS through the Genomic Data Commons portal.

The following molecular datasets were included:

\begin{itemize}
    \item DNA methylation profiles;
    \item Somatic copy-number variation (SCNV) data;
    \item Somatic mutation profiles;
    \item RNA sequencing (RNA-seq) expression data;
    \item Reverse-phase protein array (RPPA) proteomic data.
\end{itemize}
DNA methylation data were processed as CpG-site-level beta values. SCNV data were represented using segmented copy-number regions. Mutation data were binarized at the gene level according to mutation occurrence. RNA-seq data were analyzed using normalized expression values from TCGA molecular data resources. RPPA data were processed using the standardized TCGA RPPA pipeline. Samples lacking corresponding histopathology images or inferred tumor-state estimates were excluded from downstream molecular analyses. Multi-omics data were aligned with SpaTIE-derived states using TCGA patient identifiers. For statistical analyses, features with excessive missing values or extremely low variance were removed prior to downstream analysis. Detailed preprocessing procedures for each molecular modality are provided in Supplementary Methods.

\subsection{Whole-slide image preprocessing and tumor-focused patch extraction}

All WSIs were processed using OpenSdpc (\hyperlink{https://github.com/WonderLandxD/opensdpc}{https://github.com/WonderLandxD/opensdpc}) or OpenSlide (\hyperlink{https://openslide.org/}{https://openslide.org/}) under a unified preprocessing pipeline. To reduce the influence of background regions and scanning artifacts, low-resolution tissue masking was first performed to identify valid tissue foreground regions. WSIs with severe blur, tissue folding, pen marks, excessive blank regions, or substantial scanning artifacts were excluded during quality control. Because SpaTIE focuses on morphology within uterine malignancies, tumor-rich regions were preferentially retained during downstream processing.

For patch extraction, WSIs were divided into non-overlapping image patches at 20$\times$ magnification. Each patch had a spatial resolution of 224 $\times$ 224 pixels. We used a tumor-focused patch extraction procedure (Supplementary Sec. \ref{slide_check}) to build a dataset consisting of 24,318,109 patches for self-supervised pretraining. During pretraining, image augmentations included random horizontal and vertical flipping, random cropping, Gaussian blur, color jittering, and stain-related perturbation, following DINOv2-style self-supervised learning settings adapted for uterine histopathology \cite{oquab2023dinov2}. These preprocessing procedures enabled construction of a large-scale uterine histopathology corpus suitable for morphology-aware representation learning.

\subsection{Development of SpaTIE}

SpaTIE consists of two major stages: uterus-specific self-supervised representation learning and latent progression-associated tumor-state inference. In the first stage, a general-purpose pathology foundation model was adapted to uterine histopathology using parameter-efficient self-supervised learning to construct a morphology-aware latent representation space (Supplementary Sec. \ref{sm_spatie_pretraining}). In the second stage, trajectory-aware manifold inference was applied to organize static morphological representations into continuous state axes, thereby identifying morphology-derived tumor-state continua from cross-sectional histopathology (Supplementary Sec. \ref{sm_spatie_ptime}).

For uterus-specific representation learning, we adapted the universal pathology foundation model UNI \cite{chen2024uni} to the uterine pathology domain through continued self-supervised learning. UNI is based on a Vision Transformer Large architecture with 16$\times$16 patch size (ViT-L/16), comprising approximately 303 million parameters pretrained on over 100 million pathology image patches using the DINOv2 framework \cite{oquab2023dinov2, dosovitskiy2020image}. Although general pathology foundation models can capture transferable histopathological representations across tissues, many morphologic patterns in uterine malignancies are organ-specific, including glandular architecture, stromal remodeling, biphasic epithelial--mesenchymal structures, and heterogeneous tumor organization. Therefore, uterus-specific domain adaptation was performed to improve representation sensitivity to uterine tumor morphology.

To preserve general histopathological knowledge while enabling efficient uterine adaptation, we employed Low-Rank Adaptation (LoRA) \cite{hu2022lora}. Specifically, LoRA modules were inserted into the query and value projection layers of transformer attention blocks. For a pretrained weight matrix $W_0 \in \mathbb{R}^{m \times n}$, the trainable update was parameterized as:
\begin{equation}
W = W_0 + \Delta W = W_0 + BA,
\end{equation}
where $B \in \mathbb{R}^{m \times r}$ and $A \in \mathbb{R}^{r \times n}$ are trainable low-rank matrices with rank $r \ll \min(m,n)$. During training, pretrained parameters remained frozen while only LoRA parameters were optimized using the DINOv2 self-supervised objective. The model was trained for 50 epochs using the AdamW optimizer \cite{loshchilov2017decoupled} with learning rate $1e-4$, and weight decay 0.998. Model selection was based on F1-score.

After self-supervised pretraining, SpaTIE was used as a frozen feature extractor for downstream analyses. For each image patch, the CLS token embedding from the final transformer layer was extracted as the patch-level feature representation:
\begin{equation}
z_i \in \mathbb{R}^{d},
\end{equation}
where $d = 1024$ denotes the embedding dimension. To obtain slide-level representations, patch-level embeddings from the same WSI were aggregated using \cite{lu2021data}. Let $\{z_1, z_2, \dots, z_N\}$ denote patch-level embeddings from a WSI. Attention weights were computed as:
\begin{equation}
a_i = \frac{\exp\left(w^\top \tanh(Vz_i^\top)\right)}
{\sum_{j=1}^{N}\exp\left(w^\top \tanh(Vz_j^\top)\right)},
\end{equation}
where $V$ and $w$ are trainable parameters. The slide-level representation was then obtained as:
\begin{equation}
Z = \sum_{i=1}^{N} a_i z_i.
\end{equation}

To evaluate the effectiveness of uterus-specific representation learning, SpaTIE was compared with multiple baseline feature extractors, including the original universal pathology foundation model (UPFM, \cite{chen2024towards}), ImageNet-pretrained convolutional neural networks, ResNet-50-based pathology encoders, and other baseline histopathological representations. All baseline models were evaluated under identical preprocessing and downstream analysis settings whenever possible.

\subsection{Latent progression-associated tumor-state inference}

To infer latent tumor-state organization from histopathological morphology, we constructed low-dimensional manifolds from SpaTIE-derived slide-level embeddings. For each cohort, slide-level features were standardized and projected into a lower-dimensional latent space using principal component analysis (PCA), with the number of retained principal components selected to preserve at least 90\% of total variance. Neighborhood graphs were subsequently constructed using the Scanpy framework \cite{wolf2018scanpy}. Specifically, a $k$-nearest-neighbor (KNN) graph was computed based on Euclidean distances between slide-level embeddings:
\begin{equation}
G = (V,E),
\end{equation}
where each node corresponds to a WSI and edges connect neighboring samples in the latent embedding space. Affinity weights between neighboring samples were calculated using Gaussian similarity kernels:
\begin{equation}
K_{ij} = \exp\left(-\frac{\|Z_i - Z_j\|^2}{\sigma^2}\right),
\end{equation}
where $Z_i$ and $Z_j$ denote slide-level embeddings and $\sigma$ controls local neighborhood scaling. For visualization and manifold exploration, Uniform Manifold Approximation and Projection (UMAP) \cite{mcinnes2018umap} and Leiden clustering \cite{traag2019louvain} were applied to identify local morphology-associated communities within the embedding space.

Progression-associated tumor states were inferred using diffusion pseudotime (DPT) \cite{haghverdi2016diffusion}, which estimates state relationships from diffusion geometry on the manifold graph. In this study, DPT was used as a graph-ordering method for cross-sectional morphology rather than as evidence of chronological tumor evolution. Diffusion operators were computed from the neighborhood affinity matrix:
\begin{equation}
P = D^{-1}K,
\end{equation}
where $D$ is the diagonal degree matrix and $K$ is the affinity matrix. Diffusion distances were then used to estimate state relationships along the latent manifold. A root sample was selected according to low-state morphology and manifold geometry. Diffusion values were subsequently computed as geodesic distances from the root sample:
\begin{equation}
t_i = \mathrm{DPT}(Z_i),
\end{equation}
where $t_i$ denotes the latent state of sample $i$. Importantly, these inferred progression-associated states should not be interpreted as direct chronological timelines of tumor evolution. Instead, they represent morphology-derived latent state organization reconstructed from cross-sectional spatial heterogeneity.

To characterize spatial state organization within WSIs, both slide-level and patch-level progression-associated states were estimated. Patch-level embeddings were projected into the same latent manifold space, and state values were assigned using neighborhood interpolation:
\begin{equation}
t_p = f(z_p),
\end{equation}
where $z_p$ denotes a patch-level embedding and $t_p$ denotes its inferred state. Patch-level states were subsequently mapped back to original WSI coordinates to generate spatial state heatmaps. State quality was quantitatively evaluated using multiple unsupervised metrics describing latent manifold stability and spatial coherence. State stability (STAB) quantified robustness under repeated manifold reconstruction:
\begin{equation}
\mathrm{STAB} = \mathrm{corr}(t^{(1)}, t^{(2)}),
\end{equation}
where $t^{(1)}$ and $t^{(2)}$ denote states obtained from repeated analyses. Spatial coherence (SPA) measured local consistency of neighboring tissue regions:
\begin{equation}
\mathrm{SPA} = \frac{1}{N}\sum_i \frac{1}{| \mathcal{N}(i) |}
\sum_{j \in \mathcal{N}(i)}
\exp(-|t_i-t_j|),
\end{equation}
where $\mathcal{N}(i)$ denotes neighboring tissue patches. Feature smoothness (FS) quantified continuity of latent feature variation along state organization:
\begin{equation}
\mathrm{FS} = 1 - \frac{1}{N}\sum_i \|Z_i - \bar{Z}_{\mathcal{N}(i)}\|.
\end{equation}

\subsection{Static representation evaluation and downstream tasks}

To evaluate whether SpaTIE-derived static morphological representations retained clinically relevant information, multiple downstream prediction tasks were performed on TCGA-UCEC and TCGA-UCS cohorts. These tasks included histological subtype classification, tumor grade prediction, FIGO stage prediction, mutation prediction, molecular subtype prediction, and survival prediction. For each task, frozen SpaTIE-derived patch-level features were used as input to downstream MIL-based classifiers. Patient-level train-validation-test splitting was performed to avoid data leakage between slides originating from the same patient. The same downstream analysis pipeline was applied to all baseline feature extractors for fair comparison.

\subsection{Progression states evaluation metrics}
To quantitatively evaluate the quality of SpaTIE-derived static morphology representations and latent state organization, we introduced multiple unsupervised metrics describing latent manifold structure, information distribution, state robustness, spatial coherence, and feature continuity.

\paragraph{Manifold Structure Quality (MSQ).}

Manifold Structure Quality (MSQ) was designed to evaluate the intrinsic structural organization of the latent morphology manifold learned by SpaTIE. Specifically, principal component analysis (PCA) was first applied to the latent feature matrix:
\begin{equation}
X = \{z_1, z_2, \dots, z_N\}, \quad z_i \in \mathbb{R}^{d},
\end{equation}
where $z_i$ denotes the latent feature representation of the $i$-th histopathological region or WSI, $N$ denotes the number of samples, and $d$ denotes the feature dimension.

Let
\begin{equation}
\lambda_1, \lambda_2, \dots, \lambda_m
\end{equation}
denote the explained variance ratios of PCA components, where $m$ is the number of retained principal components. The intrinsic manifold dimensionality was estimated as:
\begin{equation}
\mathrm{MSQ} =
\min \left\{
k :
\sum_{i=1}^{k}\lambda_i \geq 0.9
\right\}.
\end{equation}

MSQ reflects the number of principal components required to explain 90\% of the variance in the latent space. Lower MSQ values indicate that the latent manifold is more compact and structurally organized, suggesting that morphology-associated variation is encoded within a lower-dimensional and more coherent representation space.

\paragraph{Representation Entropy (RE).}

Representation Entropy (RE) was used to evaluate the information distribution and diversity of latent morphology representations. After PCA decomposition, normalized variance coefficients were computed as:
\begin{equation}
p_i = \frac{\lambda_i}{\sum_{j=1}^{m}\lambda_j},
\end{equation}
where $\lambda_i$ denotes the variance explained by the $i$-th principal component.

The entropy of the representation space was then calculated as:
\begin{equation}
\mathrm{RE}
=
-\sum_{i=1}^{m}
p_i \log(p_i+\epsilon),
\end{equation}
where $\epsilon$ is a small constant for numerical stability.

RE measures how evenly information is distributed across latent dimensions. Higher RE values indicate richer and more diverse latent representations, whereas lower values suggest information collapse or overly concentrated latent organization.

\paragraph{Stability (STAB).}

Stability (STAB) was designed to evaluate the robustness of SpaTIE-derived state values under perturbation. Given an inferred state vector:
\begin{equation}
t = \{t_1,t_2,\dots,t_N\},
\end{equation}
where $t_i$ denotes the latent state of the $i$-th patch or WSI, Gaussian noise was added to generate a perturbed state vector:
\begin{equation}
\tilde{t}_i = t_i + \eta_i,
\quad
\eta_i \sim \mathcal{N}(0,\sigma^2).
\end{equation}

The robustness of state ordering was quantified using Spearman correlation:
\begin{equation}
\mathrm{STAB}
=
\rho_{\mathrm{Spearman}}(t,\tilde{t}),
\end{equation}
where $\rho_{\mathrm{Spearman}}$ denotes Spearman rank correlation.

Higher STAB values indicate that the inferred state organization is robust to perturbation and preserves stable relative ordering among samples.

\paragraph{Spatial Consistency (SPA).}

Spatial Consistency (SPA) was used to evaluate whether neighboring tissue regions within WSIs exhibit locally coherent state values. For each patch $i$, spatial neighbors were identified using a $k$-nearest-neighbor graph constructed from patch coordinates:
\begin{equation}
\mathcal{N}_s(i)
=
\{j_1,j_2,\dots,j_k\},
\end{equation}
where $\mathcal{N}_s(i)$ denotes the spatial neighborhood of patch $i$.

Spatial consistency was computed as:
\begin{equation}
\mathrm{SPA}
=
\frac{1}{1+
\frac{1}{N}
\sum_{i=1}^{N}
\frac{1}{|\mathcal{N}_s(i)|}
\sum_{j\in\mathcal{N}_s(i)}
|t_i-t_j|
}.
\end{equation}

SPA evaluates whether adjacent tissue regions exhibit similar state values. Higher SPA values indicate stronger local spatial coherence of state organization within tissue sections.

\paragraph{Feature Smoothness (FS).}

Feature Smoothness (FS) was designed to evaluate the continuity of latent morphology features along the inferred state axis. Let
\begin{equation}
\{z_{(1)},z_{(2)},\dots,z_{(N)}\}
\end{equation}
denote latent features sorted according to inferred state order:
\begin{equation}
t_{(1)} \leq t_{(2)} \leq \cdots \leq t_{(N)}.
\end{equation}

Feature variation between neighboring states was computed as:
\begin{equation}
\Delta_i
=
\|z_{(i+1)}-z_{(i)}\|_2.
\end{equation}

Feature smoothness was then defined as:
\begin{equation}
\mathrm{FS}
=
\frac{1}{1+\frac{1}{N-1}\sum_{i=1}^{N-1}\Delta_i}.
\end{equation}

FS measures whether latent morphology changes continuously along the inferred state organization. Higher FS values indicate smoother transitions of morphology-aware representations across inferred states, supporting the plausibility of continuous morphology-derived tumor-state organization. 

\subsection{Clinicopathological association and survival analysis}

To investigate the clinical relevance of SpaTIE-derived progression-associated tumor states, we analyzed their associations with multiple clinicopathological variables. Group-wise state differences were evaluated using Kruskal--Wallis tests followed by post hoc comparisons where appropriate. Correlations between state values and continuous clinicopathological variables were assessed using Spearman correlation analysis. For survival analysis, patients were stratified into state groups. Kaplan--Meier survival curves were generated and compared using log-rank tests. Cox proportional hazards regression was additionally performed to evaluate whether state values were associated with overall survival independently of conventional clinicopathological variables:
\begin{equation}
h(t) = h_0(t)\exp(\beta_1 x_1 + \cdots + \beta_n x_n).
\end{equation}
Hazard ratios (HRs) and 95\% confidence intervals were reported.

\subsection{Multi-omics association analysis}

To investigate the molecular correlates of SpaTIE-derived progression-associated tumor states, we integrated state values with multiple molecular modalities, including DNA methylation, somatic copy-number variation (SCNV), mutation, RNA sequencing, and RPPA proteomic data from TCGA-UCEC and TCGA-UCS. Continuous molecular features were analyzed using Spearman correlation:
\begin{equation}
\rho = \mathrm{corr}_{\mathrm{Spearman}}(x,t),
\end{equation}
where $x$ denotes the molecular feature and $t$ denotes the latent state. For categorical mutation features, state distributions between mutated and non-mutated groups were compared using Mann--Whitney U tests.

To visualize state-associated molecular variation, samples were ordered according to state values and displayed as state-ordered heatmaps. Top-ranked molecular features were identified and were subsequently used for downstream enrichment analysis. Pathway enrichment analysis was performed using Enrichr \cite{kuleshov2016enrichr} and KEGG pathway databases \cite{kanehisa2021kegg}. These analyses were designed to identify coordinated molecular programs associated with morphology-derived state organization.

\subsection{Progression-guided virtual perturbation and pathway enrichment}

To identify candidate molecular features potentially associated with state organization, we performed state-guided virtual perturbation analysis across methylation, SCNV, RNA-seq, and RPPA modalities. The concept is related to recent AI Virtual Cell frameworks, which advocate computational perturbation for biological hypothesis generation before experimental validation \cite{bunne2024build,qian2025grow}. For each candidate feature, a virtual perturbation score (KO score) was computed by evaluating the attenuation of feature--state association after computational removal of the feature-associated component:
\begin{equation}
\mathrm{KO}(g) = |\rho(t,g)| - |\rho(t_{\mathrm{perturbed}},g)|,
\end{equation}
where $g$ denotes the molecular feature, $t$ denotes the original state value, $t_{\mathrm{perturbed}}$ denotes the state value after virtual perturbation, and $\rho$ denotes the Spearman association. Features with larger KO scores were interpreted as having stronger coupling to latent state organization. Importantly, this analysis was designed as a computational hypothesis-generation strategy rather than direct causal inference or experimental knockout validation.

Top-ranked candidate genes and molecular features were further grouped into regulatory modules and subjected to pathway enrichment analysis using Enrichr and KEGG databases. These analyses were used to identify candidate biological programs potentially linked to morphology-derived states, including transcriptional regulation, chromatin remodeling, metabolic pathways, epithelial--mesenchymal organization, and signaling-related processes.

\subsection{Statistical analysis}

Continuous variables are presented as mean $\pm$ standard deviation unless otherwise specified. Group comparisons were performed using Mann--Whitney U tests or Kruskal--Wallis tests depending on the number of groups. Correlation analyses were performed using Spearman correlation coefficients. Survival analyses were conducted using Kaplan--Meier estimation, log-rank testing, and Cox proportional hazards models.

Unless otherwise specified, two-sided $p < 0.05$ was considered statistically significant. For high-dimensional molecular analyses, multiple-testing correction was evaluated using the FDR procedure where appropriate. Given the exploratory nature of the multi-omics analyses, both nominal $p$ values and FDR-adjusted statistics were reported and interpreted primarily in the context of state-associated molecular trends and candidate feature prioritization rather than definitive single-gene discovery. All statistical tests were two-sided unless otherwise specified.

\subsection{Computing hardware and software}

Model training and downstream analyses were performed on a multi-node Linux computing platform equipped with NVIDIA A800 80GB GPUs. Deep learning models were implemented using PyTorch 1.13.0 under Python 3.9.13. Whole-slide image processing was performed using OpenSlide \cite{goode2013openslide}. Manifold analysis and trajectory inference were performed using Scanpy \cite{wolf2018scanpy}, AnnData, and UMAP-related libraries. Statistical analyses were conducted using SciPy, Lifelines, Scikit-learn, and related Python scientific-computing packages.

All experiments were conducted under a unified software environment to ensure reproducibility. Source code and trained model weights will be publicly released at \hyperlink{https://github.com/jiangnansss/UterPath-SpaTIE}{https://github.com/jiangnansss/UterPath-SpaTIE}.

\section*{Data availability}
PK3-Uter is a private dataset protected by institutional ethics protocols and patient privacy regulations. Access to this dataset can be granted upon reasonable request to the corresponding authors, subject to approval by the relevant ethics committee and data use agreements. The TCGA-UCEC and TCGA-UCS datasets are publicly available through TCGA Data Portal (\hyperlink{https://portal.gdc.cancer.gov}{https://portal.gdc.cancer.gov}). The pan-cancer multi-omics data, including gene expression, copy number variation, DNA methylation, and clinical annotations, were retrieved from the UCSC Xena Browser (\hyperlink{https://xenabrowser.net/datapages/?cohort=TCGA\%20Pan-Cancer\%20(PANCAN)}{https://xenabrowser.net/datapages/?cohort=TCGA\%20Pan-Cancer\%20(PANCAN)}).

\section*{Code availability}
All custom code and analysis pipelines developed for this study are publicly available at GitHub: \hyperlink{https://github.com/jiangnansss/UterPath-SpaTIE}{https://github.com/jiangnansss/UterPath-SpaTIE}.

\section*{Supplementary information}
See details in Supplementary Materials.

\section*{Acknowledgments}
This study was supported by National Natural Science Foundation of China (82473150), National Key Research and Development Program of China (2023YFC2705805), Beijing Municipal Health Commission Research Ward Excellence Clinical Research Program (BRWEP2024WO32240114), Natural Science Foundation of Fujian Province Project (2026J01310023), Clinical Cohort Construction Program of Peking University Third Hospital (BYSYZD2022015), and Fuzhou University Talent-Introduction Scientific Startup Fund (XRC-25119).

\section*{Conflict of interest}
All authors report no disclosures.

\section*{Contributions}
Conceptualization and methodology by Qiming He, Yan Liu, Shuang Ge, Tian Guan, Congrong Liu, and Yonghong He; data curation and formal analysis by Fan Yang, Yuxiang Wang, Jing Yang, Ajin Hu, Zixiu Song, Xiaoya Zhao, Xianjing Zheng, Yijun Zheng, Liling Lin, Shuxing Liu, Bin Bao, and Yue Xie; investigation and software by Fan Yang, Ieng Man Zhang, Zihao Jia, Yexing Zhang, Qiang Huang, and Zihan Wang; resources and funding acquisition by Qiming He, Congrong Liu, and Yonghong He; writing – original draft by Qiming He, Yan Liu, and Shuang Ge; writing – review \& editing Congrong Liu, and Yonghong He; supervision by Tian Guan, Congrong Liu, and Yonghong He.

\section*{Ethics Statement and Patient Consent}
This retrospective study was approved by the Medical Ethics Committee of Peking University Third Hospital (Approval No. IRB00006761-M20250620; Ethical Review Approval Document No. (2025) Medical Ethics Review No. 577-02). The requirement for individual informed consent was waived by the Ethics Committee because the study involved the retrospective analysis of existing pathological slides and de-identified clinical data.

\bibliographystyle{MSP}
\bibliography{sample}

\newpage

\setcounter{section}{0}
\setcounter{subsection}{0}
\setcounter{figure}{0}
\setcounter{table}{0}
\setcounter{equation}{0}

\renewcommand{\thesubsection}{\arabic{subsection}}
\renewcommand{\thefigure}{S\arabic{figure}}
\renewcommand{\thetable}{S\arabic{table}}
\renewcommand{\theequation}{S\arabic{equation}}

\section*{Supplementary Material}\label{SM}
\addcontentsline{toc}{section}{Supplementary Material}

\subsection{Definition of downstream tasks}\label{sm_tasks}

The definition of downstream tasks are shown as Table \ref{sm_tasks_def}.

\begin{table*}[htbp]
\centering
\caption{Downstream prediction tasks in TCGA-UCEC and TCGA-UCS cohorts.}
\label{sm_tasks_def}

\footnotesize
\setlength{\tabcolsep}{3pt}
\renewcommand{\arraystretch}{1.15}

\begin{tabular}{
>{\centering\arraybackslash}p{2.3cm}
>{\centering\arraybackslash}p{1.3cm}
>{\centering\arraybackslash}p{1.3cm}
p{9.3cm}
}
\toprule
\textbf{Task} &
\textbf{TCGA-UCEC} &
\textbf{TCGA-UCS} &
\multicolumn{1}{c}{\textbf{Definition \& Categories}} \\
\midrule

grade & $\checkmark$ &  &
Neoplasm histologic grading (G1/G2/G3/High). \\

subtype & $\checkmark$ &  &
Molecular subtyping (CN\_HIGH/CN\_LOW/POLE/MSI). \\

FIGO & $\checkmark$ & $\checkmark$ &
FIGO stage prediction (I/II/III/IV). \\

Disease\_type & $\checkmark$ &  &
Tumor histological subtyping (Endometrioid Endometrial Adenocarcinoma, Mixed Serous and Endometrioid Carcinoma, Serous Endometrial Adenocarcinoma). \\

arid1a & $\checkmark$ &  &
ARID1A mutation status prediction (Negative/Positive). \\

kmt2d & $\checkmark$ &  &
KMT2D mutation status prediction (Negative/Positive). \\

muc16 & $\checkmark$ &  &
MUC16 mutation status prediction (Negative/Positive). \\

pik3a & $\checkmark$ & $\checkmark$ &
PIK3CA mutation status prediction (Negative/Positive). \\

pten & $\checkmark$ &  &
PTEN mutation status prediction (Negative/Positive). \\

tp53 & $\checkmark$ & $\checkmark$ &
TP53 mutation status prediction (Negative/Positive). \\

ttn & $\checkmark$ &  &
TTN mutation status prediction (Negative/Positive). \\

3\_year\_survival & $\checkmark$ & $\checkmark$ &
Three-year survival prediction (Dead/Alive). \\

5\_year\_survival & $\checkmark$ & $\checkmark$ &
Five-year survival prediction (Dead/Alive). \\

morphology & $\checkmark$ &  &
Distinguishing between UCEC and UCS (UCEC/UCS). \\

ppp2r1a &  & $\checkmark$ &
PPP2R1A mutation status prediction (Negative/Positive). \\

fbxw7 &  & $\checkmark$ &
FBXW7 mutation status prediction (Negative/Positive). \\

origin &  & $\checkmark$ &
Distinguishing between primary and metastatic UCS (Primary/Metastatic). \\

\bottomrule
\end{tabular}
\end{table*}

\subsection{Detailed results of downstream tasks}\label{sm_downstream}

The detailed results of donwstream tasks are shown in Table \ref{sm_down_1}-\ref{sm_down_6}.

\begin{table}[htbp]
\centering
\caption{Performance of all the models on TCGA-UCEC (F1-score, mean $\pm$ std). Bold indicates the optimal result for the current task.}
\label{sm_down_1}

\begin{tabular}{lccc}
\toprule
\textbf{Task} & \textbf{Baseline} & \textbf{UPFM} & \textbf{SpaTIE} \\
\midrule

disease\_type & 0.813$\pm$0.013 & \textbf{\boldmath 0.882$\pm$0.029} & 0.880$\pm$0.002 \\

subtype & 0.494$\pm$0.005 & 0.592$\pm$0.008 & \textbf{\boldmath 0.593$\pm$0.015} \\

figo & 0.352$\pm$0.010 & \textbf{\boldmath 0.453$\pm$0.007} & 0.446$\pm$0.018 \\

grade & 0.640$\pm$0.004 & \textbf{\boldmath 0.688$\pm$0.019} & 0.685$\pm$0.026 \\

pten & 0.739$\pm$0.032 & 0.842$\pm$0.015 & \textbf{\boldmath 0.846$\pm$0.017} \\

pik3a & 0.577$\pm$0.009 & \textbf{\boldmath 0.585$\pm$0.017} & 0.579$\pm$0.003 \\

arid1a & 0.623$\pm$0.007 & 0.701$\pm$0.003 & \textbf{\boldmath 0.711$\pm$0.006} \\

muc16 & 0.604$\pm$0.018 & \textbf{\boldmath 0.666$\pm$0.003} & 0.659$\pm$0.026 \\

ttn & 0.588$\pm$0.006 & 0.701$\pm$0.012 & \textbf{\boldmath 0.710$\pm$0.017} \\

kmt2d & 0.579$\pm$0.003 & \textbf{\boldmath 0.642$\pm$0.016} & 0.627$\pm$0.003 \\

tp53 & 0.722$\pm$0.010 & 0.797$\pm$0.019 & \textbf{\boldmath 0.809$\pm$0.005} \\

3-years & 0.760$\pm$0.004 & 0.758$\pm$0.027 & \textbf{\boldmath 0.765$\pm$0.002} \\

5-years & 0.674$\pm$0.015 & \textbf{\boldmath 0.742$\pm$0.007} & \textbf{\boldmath 0.742$\pm$0.002} \\

\bottomrule
\end{tabular}
\end{table}

\begin{table}[htbp]
\centering
\caption{Performance of all the models on TCGA-UCEC (Accuracy, mean $\pm$ std).}
\label{sm_down_2}
\begin{tabular}{lccc}
\toprule
\textbf{Task} & \textbf{Baseline} & \textbf{UPFM} & \textbf{SpaTIE} \\
\midrule
disease\_type & 0.860$\pm$0.003 & 0.908$\pm$0.028 & \textbf{\boldmath 0.911$\pm$0.003} \\
subtype & 0.539$\pm$0.004 & 0.629$\pm$0.022 & \textbf{\boldmath 0.630$\pm$0.011} \\
figo & 0.620$\pm$0.002 & 0.668$\pm$0.023 & \textbf{\boldmath 0.678$\pm$0.002} \\
grade & 0.706$\pm$0.008 & 0.746$\pm$0.012 & \textbf{\boldmath 0.754$\pm$0.016} \\
pten & 0.774$\pm$0.030 & 0.858$\pm$0.005 & \textbf{\boldmath 0.865$\pm$0.002} \\
pik3a & 0.577$\pm$0.009 & 0.587$\pm$0.003 & \textbf{\boldmath 0.602$\pm$0.003} \\
arid1a & 0.624$\pm$0.003 & 0.702$\pm$0.002 & \textbf{\boldmath 0.711$\pm$0.005} \\
muc16 & 0.713$\pm$0.002 & \textbf{\boldmath 0.769$\pm$0.016} & 0.749$\pm$0.003 \\
ttn & 0.611$\pm$0.005 & 0.717$\pm$0.005 & \textbf{\boldmath 0.730$\pm$0.018} \\
kmt2d & 0.698$\pm$0.002 & 0.732$\pm$0.010 & \textbf{\boldmath 0.741$\pm$0.013} \\
tp53 & 0.739$\pm$0.014 & 0.818$\pm$0.003 & \textbf{\boldmath 0.825$\pm$0.024} \\
3-years & 0.769$\pm$0.014 & 0.770$\pm$0.031 & \textbf{\boldmath 0.774$\pm$0.005} \\
5-years & 0.798$\pm$0.007 & 0.832$\pm$0.002 & \textbf{\boldmath 0.833$\pm$0.014} \\
\bottomrule
\end{tabular}
\end{table}

\begin{table}[htbp]
\centering
\caption{Performance of all the models on TCGA-UCEC (ROC-AUC, mean $\pm$ std).}
\label{sm_down_3}
\begin{tabular}{lccc}
\toprule
\textbf{Task} & \textbf{Baseline} & \textbf{UPFM} & \textbf{SpaTIE} \\
\midrule
disease\_type & 0.893$\pm$0.026 & 0.934$\pm$0.029 & \textbf{\boldmath 0.938$\pm$0.025} \\
subtype & 0.729$\pm$0.008 & 0.798$\pm$0.005 & \textbf{\boldmath 0.799$\pm$0.014} \\
figo & 0.613$\pm$0.002 & 0.684$\pm$0.007 & \textbf{\boldmath 0.693$\pm$0.004} \\
grade & 0.837$\pm$0.008 & \textbf{\boldmath 0.866$\pm$0.030} & 0.865$\pm$0.009 \\
pten & 0.801$\pm$0.027 & \textbf{\boldmath 0.890$\pm$0.002} & 0.889$\pm$0.006 \\
pik3a & 0.587$\pm$0.004 & 0.592$\pm$0.022 & \textbf{\boldmath 0.602$\pm$0.005} \\
arid1a & 0.651$\pm$0.007 & 0.748$\pm$0.005 & \textbf{\boldmath 0.751$\pm$0.007} \\
muc16 & 0.633$\pm$0.028 & 0.704$\pm$0.024 & \textbf{\boldmath 0.736$\pm$0.008} \\
ttn & 0.596$\pm$0.008 & \textbf{\boldmath 0.705$\pm$0.009} & \textbf{\boldmath 0.705$\pm$0.009} \\
kmt2d & 0.633$\pm$0.004 & 0.670$\pm$0.006 & \textbf{\boldmath 0.683$\pm$0.005} \\
tp53 & 0.776$\pm$0.003 & 0.837$\pm$0.016 & \textbf{\boldmath 0.844$\pm$0.015} \\
3-years & 0.805$\pm$0.004 & 0.821$\pm$0.004 & \textbf{\boldmath 0.824$\pm$0.026} \\
5-years & 0.776$\pm$0.018 & 0.845$\pm$0.024 & \textbf{\boldmath 0.846$\pm$0.024} \\
\bottomrule
\end{tabular}
\end{table}

\begin{table}[htbp]
\centering
\caption{Performance of all the models on TCGA-UCS (F1-score, mean $\pm$ std).}
\label{sm_down_4}
\begin{tabular}{lccc}
\toprule
\textbf{Task} & \textbf{Baseline} & \textbf{UPFM} & \textbf{SpaTIE} \\
\midrule
morphology & 0.844$\pm$0.002 & 0.788$\pm$0.002 & \textbf{\boldmath 0.854$\pm$0.020} \\
origin & 0.750$\pm$0.010 & 0.763$\pm$0.028 & \textbf{\boldmath 0.800$\pm$0.005} \\
figo & \textbf{\boldmath 0.589$\pm$0.019} & 0.523$\pm$0.005 & 0.559$\pm$0.002 \\
fbxw7 & 0.711$\pm$0.006 & 0.665$\pm$0.002 & \textbf{\boldmath 0.735$\pm$0.007} \\
pik3a & 0.676$\pm$0.031 & 0.743$\pm$0.007 & \textbf{\boldmath 0.746$\pm$0.004} \\
ppp2r1a & 0.579$\pm$0.012 & \textbf{\boldmath 0.642$\pm$0.007} & 0.627$\pm$0.007 \\
tp53 & 0.768$\pm$0.002 & 0.838$\pm$0.004 & \textbf{\boldmath 0.840$\pm$0.017} \\
\bottomrule
\end{tabular}
\end{table}

\begin{table}[htbp]
\centering
\caption{Performance of all the models on TCGA-UCS (Accuracy, mean $\pm$ std).}
\label{sm_down_5}
\begin{tabular}{lccc}
\toprule
\textbf{Task} & \textbf{Baseline} & \textbf{UPFM} & \textbf{SpaTIE} \\
\midrule
morphology & 0.900$\pm$0.004 & 0.880$\pm$0.009 & \textbf{\boldmath 0.926$\pm$0.003} \\
origin & 0.781$\pm$0.005 & 0.792$\pm$0.006 & \textbf{\boldmath 0.825$\pm$0.012} \\
figo & 0.605$\pm$0.024 & 0.594$\pm$0.003 & \textbf{\boldmath 0.627$\pm$0.028} \\
fbxw7 & 0.724$\pm$0.021 & 0.756$\pm$0.002 & \textbf{\boldmath 0.759$\pm$0.009} \\
pik3a & 0.726$\pm$0.014 & 0.759$\pm$0.008 & \textbf{\boldmath 0.767$\pm$0.018} \\
ppp2r1a & 0.698$\pm$0.025 & 0.732$\pm$0.008 & \textbf{\boldmath 0.741$\pm$0.013} \\
tp53 & 0.923$\pm$0.021 & 0.944$\pm$0.010 & \textbf{\boldmath 0.945$\pm$0.002} \\
\bottomrule
\end{tabular}
\end{table}

\begin{table}[htbp]
\centering
\caption{Performance of all the models on TCGA-UCS (ROC-AUC, mean $\pm$ std).}
\label{sm_down_6}
\begin{tabular}{lccc}
\toprule
\textbf{Task} & \textbf{Baseline} & \textbf{UPFM} & \textbf{SpaTIE} \\
\midrule
morphology & 0.924$\pm$0.030 & \textbf{\boldmath 0.935$\pm$0.005} & 0.905$\pm$0.025 \\
origin & 0.783$\pm$0.019 & \textbf{\boldmath 0.822$\pm$0.004} & 0.821$\pm$0.027 \\
figo & 0.734$\pm$0.025 & 0.710$\pm$0.013 & \textbf{\boldmath 0.712$\pm$0.010} \\
fbxw7 & 0.698$\pm$0.026 & \textbf{\boldmath 0.730$\pm$0.002} & 0.713$\pm$0.004 \\
pik3a & 0.708$\pm$0.005 & 0.780$\pm$0.019 & \textbf{\boldmath 0.823$\pm$0.023} \\
ppp2r1a & 0.633$\pm$0.007 & 0.670$\pm$0.006 & \textbf{\boldmath 0.683$\pm$0.026} \\
tp53 & 0.818$\pm$0.004 & \textbf{\boldmath 0.952$\pm$0.005} & 0.950$\pm$0.013 \\
\bottomrule
\end{tabular}
\end{table}

\subsection{Tumor-focused patch extraction}\label{slide_check}

To improve pretraining efficiency and reduce background-induced noise, we performed a tumor-focused patch extraction procedure before self-supervised pretraining. Directly extracting patches from all tissue regions yielded more than 60 million candidate patches, many of which corresponded to normal tissue, stromal regions, or low-information areas unrelated to tumor morphology. To concentrate the pretraining corpus on diagnostically relevant regions, we additionally annotated 10,000 tumor patches and 10,000 non-tumor patches from the uterine pathology slides and trained a binary tumor-region classifier.

Previous tumor region segmentation methods have provided insights into the extraction of key patches \cite{he2023unsupervised,liu2022using}. Specifically, we initialized a ViT-L backbone with UNI weights \cite{chen2024uni} and fine-tuned the model for tumor-versus-non-tumor classification using standard cross-entropy loss. The trained classifier was then applied to the full patch pool, and only patches predicted as tumor regions were retained for subsequent self-supervised pretraining. This procedure reduced the number of pretraining patches from over 60 million to approximately 20 million, substantially increasing the tumor-content ratio and improving the signal-to-noise characteristics of the pretraining dataset.

\begin{table}[htbp]
\centering
\caption{Associations between SpaTIE-derived state values and clinicopathological variables in TCGA-UCEC and TCGA-UCS cohorts.}\label{sm_ptime_clinic}
\label{tab:clinical_association}
\begin{tabular}{llccc}
\hline
Cohort & Clinical variable & Statistical value & $p$ value & FDR \\
\hline

\multicolumn{5}{l}{\textbf{TCGA-UCEC}}\\
\hline

UCEC & Fraction Genome Altered (FGA) 
& $\rho=-0.181$ 
& $1.76\times10^{-5}$ 
& $5.29\times10^{-5}$ \\

UCEC & Diagnosis Age 
& $\rho=-0.071$ 
& 0.0930 
& 0.1507 \\

UCEC & Birth from Initial Pathologic Diagnosis Date 
& $\rho=0.069$ 
& 0.1004 
& 0.1507 \\

UCEC & TMB (nonsynonymous) 
& $\rho=-0.051$ 
& 0.2433 
& 0.2920 \\

UCEC & Mutation Count 
& $\rho=-0.041$ 
& 0.3442 
& 0.3442 \\

UCEC & FIGO Major Stage 
& $H=15.357$ 
& 0.0015 
& 0.0077 \\

UCEC & Prior Malignancy 
& $U=13936.0$ 
& 0.0078 
& 0.0259 \\

UCEC & Patient Vital Status 
& $U=23647.0$ 
& 0.0262 
& 0.0656 \\

UCEC & Ethnicity Category 
& $U=2334.0$ 
& 0.0355 
& 0.0709 \\

UCEC & Race Category 
& $H=8.671$ 
& 0.0699 
& 0.1164 \\

UCEC & Morphology 
& $H=5.093$ 
& 0.1651 
& 0.2064 \\

UCEC & Disease Type 
& $H=2.700$ 
& 0.2593 
& 0.2593 \\

UCEC & Primary Diagnosis 
& $H=2.700$ 
& 0.2593 
& 0.2593 \\

\hline

\multicolumn{5}{l}{\textbf{TCGA-UCS}}\\
\hline

UCS & Year of Diagnosis 
& $\rho=-0.241$ 
& 0.0215 
& 0.1130 \\

UCS & Fraction Genome Altered (FGA) 
& $\rho=0.219$ 
& 0.0377 
& 0.1130 \\

UCS & Mutation Count 
& $\rho=0.133$ 
& 0.2091 
& 0.3015 \\

UCS & TMB (nonsynonymous) 
& $\rho=0.133$ 
& 0.2100 
& 0.3015 \\

UCS & Birth from Initial Pathologic Diagnosis Date 
& $\rho=0.121$ 
& 0.2533 
& 0.3015 \\

UCS & Diagnosis Age 
& $\rho=-0.109$ 
& 0.3015 
& 0.3015 \\

UCS & Morphology 
& $U=916.0$ 
& 0.0047 
& 0.0379 \\

UCS & Race Category 
& $H=8.847$ 
& 0.0120 
& 0.0480 \\

UCS & Disease Type 
& $H=2.706$ 
& 0.2584 
& 0.4669 \\

UCS & Primary Diagnosis 
& $H=2.706$ 
& 0.2584 
& 0.4669 \\

UCS & FIGO Major Stage 
& $H=2.795$ 
& 0.4243 
& 0.5657 \\

UCS & Prior Malignancy 
& $U=515.0$ 
& 0.6347 
& 0.7254 \\

UCS & Patient Vital Status 
& $U=991.0$ 
& 0.8600 
& 0.8600 \\

\hline
\end{tabular}
\end{table}

\subsection{Details of association between SpaTIE-derived progression states and clinical features}

The details of association between SpaTIE-derived progression states and clinical features are shown in Table \ref{sm_ptime_clinic}.

\subsection{Details of association between SpaTIE-derived progression states and multi-omics tumor programs}\label{sm_simple_association}

\begin{table}[htbp]
\centering
\caption{Pathway-level associations between SpaTIE-derived progression states and multi-omics pathway scores. Pathway scores were compared between low- and high-state groups. These results are used as pathway-level context and should not be interpreted as evidence that all member features are individually significant.}
\label{sm_pathway_assoc}
\begin{tabular}{lllccc}
\hline
Cohort & Omics & Pathway & $p$ value & $n$ & Higher group \\
\hline
UCS  & RNASeq      & Estrogen\_Response & 0.0048 & 57  & low-state \\
UCEC & Methylation & PRC2\_Polycomb     & 0.0189 & 389 & high-state \\
UCEC & SCNV        & MMR\_CNV           & 0.0222 & 496 & low-state \\
UCEC & RPPA        & AKT\_mTOR          & 0.0249 & 405 & low-state \\
UCEC & RPPA        & Cell\_Cycle        & 0.0310 & 405 & low-state \\
\hline
\end{tabular}
\end{table}

We re-evaluated the association between SpaTIE-derived progression states and multi-omics profiles using DNA methylation, somatic copy-number variation (SCNV), mutation, RNA sequencing (RNASeq), and RPPA data from TCGA-UCEC and TCGA-UCS. Because this analysis involved many molecular features and the UCS cohort was comparatively small, we emphasized signals that were supported after multiple-testing correction or were reproduced at pathway level. Nominal associations with FDR $>$ 0.05 are reported only as exploratory observations.

At the pathway level, only a limited number of pathway-score differences were detected between low- and high-state groups (Table \ref{sm_pathway_assoc}). In TCGA-UCEC, significant nominal differences were observed for methylation-based PRC2/Polycomb scores, SCNV-based MMR-related scores, and RPPA-based AKT--mTOR and cell-cycle scores. In TCGA-UCS, RNASeq-based estrogen-response scores differed between state groups. These pathway-level findings support the interpretation that the morphology-derived state axis is associated with selected molecular programs rather than with a broad, uniformly significant multi-omics shift.

Feature-level analyses showed the clearest corrected support in TCGA-UCEC. The strongest FDR-supported signals were concentrated in DNA methylation, SCNV, and RPPA layers, whereas mutation and RNASeq features were mainly nominal in the current analysis. Representative UCEC features that were also prioritized by virtual perturbation included methylation features such as \textit{PRR22}, \textit{RBM27}, and \textit{LRRC42}; SCNV features such as \textit{FAM81A}, \textit{RNA5SP396}, and \textit{GCNT3}; and RPPA features such as AXL, GAPDH, ERRFI1, CTNNA1, and SERPINE1 (Table \ref{sm_ko_overview}). These results suggest that the UCEC state axis has measurable molecular correlates, especially in epigenetic, copy-number, and protein-level measurements.

In TCGA-UCS, several individual features showed relatively large effect sizes, including RNASeq, methylation, RPPA, mutation, and cytoband-level SCNV candidates. However, the top UCS feature-level associations did not show FDR support in the current sample. Therefore, UCS molecular results are retained as exploratory candidates for hypothesis generation, not as validated molecular correlates of the morphology-derived state axis. Overall, the multi-omics analyses support a cautious interpretation: SpaTIE-derived states are accompanied by selected molecular variation, with stronger statistical support in UCEC than in UCS.

\subsection{Details of progression-guided virtual perturbation analysis}\label{sm_KO}

Progression-guided virtual perturbation was used as a computational feature-prioritization procedure. For each molecular feature, we estimated how much the feature--state association was attenuated after removing the feature-associated component from the inferred state values. The resulting score should not be interpreted as an experimental knockout effect or as causal evidence. It is better viewed as a ranking statistic that highlights features whose measured variation is tightly coupled to the morphology-derived state axis.

The updated perturbation results are summarized in Table \ref{sm_ko_overview}. In TCGA-UCEC, the top-ranked features were \textit{PRR22} for methylation, \textit{GPR84} for mutation, \textit{FAM81A} for SCNV, \textit{RAB27B} for RNASeq, and AXL for RPPA. Among the top 20 perturbation hits, FDR-supported state associations were present for methylation (20/20), SCNV (3/20), and RPPA (5/20), but not for mutation or RNASeq. Accordingly, we place more weight on UCEC methylation, SCNV, and RPPA findings, and treat UCEC mutation and RNASeq candidates as exploratory.

In TCGA-UCS, the top-ranked perturbation features were \textit{PIK3C2G} for methylation, \textit{TPTE} for mutation, 16q23.1 for SCNV, \textit{UFD1L} for RNASeq, and ARHI for RPPA. Although the perturbation scores were sometimes larger than those observed in UCEC, none of the top 20 features in the UCS modalities had feature-level FDR support. This pattern is consistent with limited statistical power and greater uncertainty in the UCS cohort. We therefore describe UCS perturbation results as exploratory molecular leads.

\begin{table}[htbp]
\centering
\caption{Overview of progression-guided virtual perturbation results. The last column reports how many of the top 20 perturbation-ranked features also showed FDR-supported association with SpaTIE-derived progression states in the feature-level analysis.}
\label{sm_ko_overview}
\begin{tabular}{lllcccc}
\hline
Cohort & Omics & Top feature & KO score & Base $\rho$ & $p$ value & FDR-supported top 20 \\
\hline
UCEC & Methylation & PRR22    & 0.312 &  0.338 & $7.75\times10^{-12}$ & 20 \\
UCEC & Mutation    & GPR84    & 2.307 & -0.176 & $5.81\times10^{-3}$  & 0 \\
UCEC & SCNV        & FAM81A   & 0.138 &  0.144 & $1.28\times10^{-3}$  & 3 \\
UCEC & RNASeq      & RAB27B   & 0.302 & -0.302 & $1.70\times10^{-4}$  & 0 \\
UCEC & RPPA        & AXL      & 0.186 &  0.189 & $1.25\times10^{-4}$  & 5 \\
UCS  & Methylation & PIK3C2G  & 0.428 & -0.465 & $2.64\times10^{-4}$  & 0 \\
UCS  & Mutation    & TPTE     & 1.546 & -0.339 & $9.88\times10^{-3}$  & 0 \\
UCS  & SCNV        & 16q23.1  & 0.274 & -0.297 & $2.62\times10^{-2}$  & 0 \\
UCS  & RNASeq      & UFD1L    & 0.443 &  0.492 & $1.01\times10^{-4}$  & 0 \\
UCS  & RPPA        & ARHI     & 0.385 &  0.409 & $3.92\times10^{-3}$  & 0 \\
\hline
\end{tabular}
\end{table}

Cross-omics inspection did not reveal a robust same-gene chain spanning DNA, RNA, and protein layers. In UCEC, the most consistent support came from multiple FDR-supported methylation, SCNV, and RPPA candidates rather than from a single linear molecular cascade. In UCS, two same-gene methylation--RNA candidates, \textit{MGC87042} and \textit{ROR2}, appeared among the top 20 perturbation hits across layers, but neither had feature-level FDR support. Several cross-layer correlations were observed among top-ranked candidates, especially between methylation or mutation and RNASeq features in UCS, and between RNASeq and RPPA features in UCEC. Because these edges are based on bulk Spearman correlations, we interpret them only as candidate inter-layer relationships.

Taken together, virtual perturbation provides a ranked set of molecular hypotheses linked to SpaTIE-derived state organization. The most statistically supported evidence in the current data comes from UCEC methylation, SCNV, and RPPA features. UCS perturbation results and all features without FDR support should be considered exploratory and require independent validation or experimental follow-up.

\subsection{Detailed Pretraining Methodology for SpaTIE}\label{sm_spatie_pretraining}

\subsubsection{Model initialization}

We selected UNI~\cite{chen2024towards} as the initialization for our domain-specific pretraining. UNI is a general-purpose computational pathology foundation model built upon the Vision Transformer with Large architecture and $16\times16$ patch size (ViT-L/16). It comprises approximately 303 million parameters and was pretrained on the Mass-100K dataset, which contains over 100 million tissue patches extracted from 100,426 diagnostic WSIs across 20 major organ types, encompassing normal tissue, cancerous tissue, and other pathological conditions~\cite{chen2024towards}. The pretraining of UNI employed the DINOv2 self-supervised learning framework~\cite{oquab2023dinov2}, which leverages a student-teacher distillation paradigm to learn robust visual representations without manual annotations. By initializing from UNI, SpaTIE inherits general histopathological feature extraction capabilities that have been demonstrated to generalize across 33 diverse anatomic pathology tasks.

\subsubsection{Vision Transformer Architecture and Self-Attention Mechanism}

SpaTIE retains the ViT-L/16 architecture from UNI. The model processes an input image $\mathbf{x} \in \mathbb{R}^{H \times W \times C}$ by dividing it into a sequence of flattened non-overlapping patches $\mathbf{x}_p \in \mathbb{R}^{N \times (P^2 \cdot C)}$, where $P=16$ is the patch size, $C=3$ denotes the RGB channels, and $N = HW/P^2$ is the total number of patches. Each patch is linearly projected via a trainable matrix $\mathbf{E} \in \mathbb{R}^{(P^2 \cdot C) \times D}$ to obtain patch embeddings, where $D=1024$ is the embedding dimension for ViT-L. A learnable class token $\mathbf{x}_{\mathrm{cls}}$ and sinusoidal positional embeddings $\mathbf{E}_{\mathrm{pos}} \in \mathbb{R}^{(N+1) \times D}$ are added to form the input sequence $\mathbf{z}_0 = [\mathbf{x}_{\mathrm{cls}}; \mathbf{x}_p^1\mathbf{E}; \dots; \mathbf{x}_p^N\mathbf{E}] + \mathbf{E}_{\mathrm{pos}}$.

The sequence $\mathbf{z}_0$ is fed through $L=24$ identical transformer encoder blocks. Each block $l$ consists of Layer Normalization (LN), multi-head self-attention (MSA), and a feed-forward network (FFN), with residual connections:

\begin{align}
\mathbf{z}'_l &= \mathrm{MSA}(\mathrm{LN}(\mathbf{z}_{l-1})) + \mathbf{z}_{l-1}, \\
\mathbf{z}_l &= \mathrm{FFN}(\mathrm{LN}(\mathbf{z}'_l)) + \mathbf{z}'_l.
\end{align}

The multi-head self-attention mechanism is the core computational primitive of the transformer. For an input sequence $\mathbf{Z} \in \mathbb{R}^{(N+1) \times D}$, the model first projects it into query, key, and value matrices:

\begin{equation}
\mathbf{Q} = \mathbf{Z}\mathbf{W}^Q, \quad \mathbf{K} = \mathbf{Z}\mathbf{W}^K, \quad \mathbf{V} = \mathbf{Z}\mathbf{W}^V,
\end{equation}
where $\mathbf{W}^Q, \mathbf{W}^K, \mathbf{W}^V \in \mathbb{R}^{D \times D}$ are learnable projection matrices. For multi-head attention with $h$ heads, these projections are partitioned into $h$ groups, each with dimension $d_k = D/h$. The scaled dot-product attention for each head $i$ is computed as:

\begin{equation}
\mathrm{head}_i = \mathrm{Attention}(\mathbf{Q}_i, \mathbf{K}_i, \mathbf{V}_i) = \mathrm{softmax}\left(\frac{\mathbf{Q}_i\mathbf{K}_i^\top}{\sqrt{d_k}}\right)\mathbf{V}_i.
\end{equation}
The outputs of all heads are concatenated and linearly projected:

\begin{equation}
\mathrm{MSA}(\mathbf{Z}) = \mathrm{Concat}(\mathrm{head}_1, \dots, \mathrm{head}_h)\mathbf{W}^O,
\end{equation}
where $\mathbf{W}^O \in \mathbb{R}^{D \times D}$. The self-attention mechanism enables each patch token to attend to all other tokens, capturing long-range spatial dependencies and contextual relationships critical for histopathological pattern recognition.

\subsubsection{Parameter-Efficient Fine-Tuning with LoRA}

To adapt the pretrained UNI model to the uterine pathology domain without catastrophic forgetting of general histopathological knowledge, we employed Low-Rank Adaptation (LoRA)~\cite{hu2022lora}, a parameter-efficient fine-tuning (PEFT) technique. LoRA is grounded in the hypothesis that the intrinsic dimensionality of fine-tuning adaptations is low, and thus the weight updates can be effectively approximated by low-rank matrices.

For each linear layer with pretrained weight matrix $\mathbf{W}_0 \in \mathbb{R}^{m \times n}$ in the transformer, LoRA introduces a trainable low-rank decomposition to approximate the full fine-tuning update $\Delta\mathbf{W}$:

\begin{equation}
\mathbf{W} = \mathbf{W}_0 + \Delta\mathbf{W} = \mathbf{W}_0 + \mathbf{B}\mathbf{A},
\end{equation}
where $\mathbf{B} \in \mathbb{R}^{m \times r}$ and $\mathbf{A} \in \mathbb{R}^{r \times n}$ are the trainable low-rank matrices, and the rank $r \ll \min(m,n)$ is a hyperparameter. During training, the original pretrained weights $\mathbf{W}_0$ remain frozen, and only $\mathbf{A}$ and $\mathbf{B}$ are updated via gradient descent. This reduces the number of trainable parameters from $mn$ to $r(m+n)$, achieving substantial memory and computational savings. Following standard practice~\cite{hu2022lora}, we applied LoRA to the query ($\mathbf{W}^Q$) and value ($\mathbf{W}^V$) projection matrices in all $L=24$ self-attention layers. The matrices $\mathbf{A}$ were initialized with random Gaussian values and $\mathbf{B}$ with zeros, such that $\Delta\mathbf{W} = \mathbf{B}\mathbf{A} = \mathbf{0}$ at initialization, ensuring that the adaptation starts from the pretrained UNI representation.

\subsubsection{DINOv2 Self-Supervised Pretraining Objective}

We continued pretraining the LoRA-adapted UNI model using the DINOv2 self-supervised learning framework~\cite{oquab2023dinov2}, which extends the original DINO approach with improved training stability and representation quality. DINOv2 employs a student-teacher self-distillation architecture where the student network processes augmented views of an input image and learns to predict the output distribution of a momentum-updated teacher network.

\paragraph{Teacher-Student Architecture.} Let $\theta_s$ and $\theta_t$ denote the parameters of the student and teacher networks, respectively. The student network is updated via backpropagation, while the teacher parameters are computed as an exponential moving average (EMA) of the student parameters:

\begin{equation}
\theta_t \leftarrow \lambda\theta_t + (1-\lambda)\theta_s,
\end{equation}
where $\lambda \in [0,1)$ is the EMA decay rate. Both networks share the same ViT-L/16 architecture with LoRA-adapted attention layers. During training, multiple augmented views are generated from each image: a set of global views (covering a large portion of the image) and local views (covering smaller regions). Only the global views are passed through the teacher network, while all views are processed by the student network.

\paragraph{DINO Loss (Image-Level Self-Distillation).} The DINO loss enforces consistency between the student and teacher outputs at the image level using the class token ($\mathrm{cls}$) representations. Let $g_s(\mathbf{x})$ and $g_t(\mathbf{x})$ denote the output logits of the student and teacher, respectively, obtained by projecting the $\mathrm{cls}$ token features through a learnable projection head followed by a softmax normalization with temperature scaling. The DINO loss is formulated as the cross-entropy between the teacher's target distribution and the student's predicted distribution:

\begin{equation}
\mathcal{L}_{\mathrm{DINO}} = -\sum_{c=1}^{C} p_t^c(\mathbf{x}) \log p_s^c(\mathbf{x}),
\end{equation}
where $p_s^c(\mathbf{x}) = \frac{\exp(g_s^c(\mathbf{x})/\tau_s)}{\sum_{j=1}^{C}\exp(g_s^j(\mathbf{x})/\tau_s)}$ and $p_t^c(\mathbf{x}) = \frac{\exp(g_t^c(\mathbf{x})/\tau_t)}{\sum_{j=1}^{C}\exp(g_t^j(\mathbf{x})/\tau_t)}$ are the probability distributions of the student and teacher, respectively, $C$ is the dimension of the projection head output, and $\tau_s$, $\tau_t$ are temperature parameters. The teacher outputs are typically centered with a running mean to prevent collapse.

\paragraph{iBOT Loss (Patch-Level Masked Modeling).} DINOv2 integrates the iBOT objective to enhance spatial consistency and local feature learning. Random patches in the student input are masked, and the student is trained to predict the teacher's output distribution for these masked patches using their unmasked context. Let $\mathcal{M}$ denote the set of masked patch indices. For each masked patch $i \in \mathcal{M}$, the iBOT loss is:

\begin{equation}
\mathcal{L}_{\mathrm{iBOT}} = -\sum_{i \in \mathcal{M}} \sum_{c=1}^{C} p_t^{c,i}(\mathbf{x}) \log p_s^{c,i}(\mathbf{x}),
\end{equation}
where $p_s^{c,i}$ and $p_t^{c,i}$ are the student and teacher probability distributions for patch $i$, respectively.

\paragraph{KoLeo Loss.} DINOv2 additionally employs the KoLeo divergence loss to encourage uniformity in the representation space and prevent representation collapse. For a batch of $B$ samples with normalized features $\{\mathbf{f}_i\}_{i=1}^B$, the KoLeo loss is:

\begin{equation}
\mathcal{L}_{\mathrm{KoLeo}} = -\frac{1}{B}\sum_{i=1}^{B} \log \min_{j \neq i} \|\mathbf{f}_i - \mathbf{f}_j\|_2.
\end{equation}

\paragraph{Total Training Objective.} The overall DINOv2 loss combines the above components:

\begin{equation}
\mathcal{L}_{\mathrm{DINOv2}} = \mathcal{L}_{\mathrm{DINO}} + \lambda_{\mathrm{iBOT}}\mathcal{L}_{\mathrm{iBOT}} + \lambda_{\mathrm{KoLeo}}\mathcal{L}_{\mathrm{KoLeo}},
\end{equation}
where $\lambda_{\mathrm{iBOT}}$ and $\lambda_{\mathrm{KoLeo}}$ are weighting hyperparameters. Through continued pretraining with this objective on uterine pathology data, the LoRA-adapted parameters progressively shift the model's feature space toward uterine-specific morphological patterns while the frozen UNI backbone preserves general histopathological priors, yielding the uterine pathology foundation model SpaTIE.

\subsection{Detailed Methodology for Pseudo-temporal Mapping}\label{sm_spatie_ptime}

\subsubsection{Feature Sampling and Dimensionality Reduction}

Pseudo-temporal analysis was conducted on slide-level features extracted by SpaTIE. To ensure computational tractability while preserving population diversity, we randomly subsampled $1\%$ of the full cohort without replacement, yielding a representative subset of slides. Each slide was represented by a $D=1024$-dimensional feature vector $\mathbf{f}_i \in \mathbb{R}^{1024}$ obtained from the frozen SpaTIE backbone. The feature matrix $\mathbf{X} \in \mathbb{R}^{N \times 1024}$ was assembled by stacking all subsampled slide features, where $N$ is the number of selected slides.

Due to the high dimensionality and potential redundancy in the feature space, we applied principal component analysis (PCA) as a preprocessing step. PCA seeks an orthogonal linear transformation that maps the data to a new coordinate system such that the greatest variance lies on the first coordinate (principal component), the second greatest variance on the second coordinate, and so on. Formally, given the centered data matrix $\tilde{\mathbf{X}} = \mathbf{X} - \bar{\mathbf{X}}$, we solve the eigenvalue decomposition of the covariance matrix:

\begin{equation}
\mathbf{C} = \frac{1}{N-1}\tilde{\mathbf{X}}^\top\tilde{\mathbf{X}} = \mathbf{U}\boldsymbol{\Lambda}\mathbf{U}^\top,
\end{equation}
where $\mathbf{U} \in \mathbb{R}^{1024 \times 1024}$ contains the orthonormal eigenvectors and $\boldsymbol{\Lambda} = \mathrm{diag}(\lambda_1, \dots, \lambda_{1024})$ with $\lambda_1 \geq \lambda_2 \geq \dots \geq \lambda_{1024} \geq 0$ are the eigenvalues representing the variance explained by each component. Rather than fixing the number of components a priori, we retained the minimal set of principal components explaining $90\%$ of the cumulative variance:

\begin{equation}
d = \min\left\{ k : \frac{\sum_{i=1}^{k}\lambda_i}{\sum_{i=1}^{1024}\lambda_i} \geq 0.9 \right\}.
\end{equation}
The projected data $\mathbf{X}_{\mathrm{pca}} = \tilde{\mathbf{X}}\mathbf{U}_{:,1:d} \in \mathbb{R}^{N \times d}$ served as the input for subsequent graph-based trajectory inference.

\subsubsection{K-Nearest Neighbor Graph Construction}

To capture the local manifold structure of the histopathological state space, we constructed an undirected $k$-nearest neighbor ($k$-NN) graph. For each slide $i$, we identified its $k=10$ nearest neighbors $\mathcal{N}_k(i)$ based on Euclidean distance in the PCA-reduced space. The adjacency matrix $\mathbf{A} \in \mathbb{R}^{N \times N}$ was defined as:

\begin{equation}
A_{ij} = \begin{cases}
1 & \text{if } j \in \mathcal{N}_k(i) \text{ or } i \in \mathcal{N}_k(j), \\
0 & \text{otherwise}.
\end{cases}
\end{equation}
The degree matrix $\mathbf{D} = \mathrm{diag}(d_1, \dots, d_N)$ with $d_i = \sum_j A_{ij}$ was computed for subsequent graph operations.

\subsubsection{Leiden Community Detection}

To identify biologically coherent histopathological subgroups, we applied the Leiden algorithm for community detection on the $k$-NN graph. The Leiden algorithm optimizes the modularity-like quality function known as the Constant Potts Model (CPM), which for resolution parameter $\gamma$ is defined as:

\begin{equation}
\mathcal{H}(\sigma) = -\sum_{ij}\left(A_{ij} - \gamma\right)\delta(\sigma_i, \sigma_j),
\end{equation}
where $\sigma_i \in \{1, \dots, K\}$ denotes the cluster assignment of node $i$, and $\delta(\cdot, \cdot)$ is the Kronecker delta. The algorithm proceeds through three phases: (1) local moving of nodes to maximize $\mathcal{H}(\sigma)$; (2) refinement of the partition to guarantee that all clusters are connected and well-separated; and (3) aggregation of the network based on the refined partition. We set $\gamma=1.0$ to obtain a moderate number of clusters balancing granularity and statistical power. The resulting partition $\mathcal{P} = \{C_1, \dots, C_K\}$ was stored as categorical annotations in the AnnData object.

\subsubsection{Partition-Based Graph Abstraction (PAGA)}

To infer a coarse-grained representation of the developmental trajectory while preserving biological meaningfulness, we employed PAGA. PAGA constructs a graph where each node represents a Leiden cluster and edges encode the statistical significance of inter-cluster connectivity. For two clusters $C_\alpha$ and $C_\beta$, the connectivity score is computed as:

\begin{equation}
w_{\alpha\beta} = \frac{\sum_{i \in C_\alpha, j \in C_\beta} A_{ij} + \sum_{i \in C_\beta, j \in C_\alpha} A_{ij}}{|C_\alpha| \cdot |C_\beta|},
\end{equation}
normalized by the expected connectivity under a random null model. The statistical significance is assessed via a one-sample test comparing observed to expected connectivities. The PAGA graph $\mathcal{G}_{\mathrm{PAGA}} = (\mathcal{V}_{\mathrm{PAGA}}, \mathcal{E}_{\mathrm{PAGA}})$ provides a topologically meaningful backbone for trajectory inference, with edges weighted by $-\log_{10}(p_{\alpha\beta})$, where $p_{\alpha\beta}$ is the $p$-value of the connectivity test.

\subsubsection{Diffusion Pseudotime (DPT) Computation}

To establish a quantitative ordering of histopathological progression, we computed diffusion pseudotime using the random-walk normalized graph Laplacian. The random-walk Laplacian is defined as:

\begin{equation}
\mathbf{L}_{\mathrm{rw}} = \mathbf{I} - \mathbf{D}^{-1}\mathbf{A},
\end{equation}
where $\mathbf{I}$ is the identity matrix and $\mathbf{D}^{-1}\mathbf{A}$ is the random-walk transition matrix. The eigendecomposition of $\mathbf{L}_{\mathrm{rw}}$ yields:

\begin{equation}
\mathbf{L}_{\mathrm{rw}} \boldsymbol{\psi}_i = \lambda_i \boldsymbol{\psi}_i, \quad i = 1, \dots, N,
\end{equation}
with eigenvalues $0 = \lambda_1 < \lambda_2 \leq \dots \leq \lambda_N$ and corresponding eigenvectors $\boldsymbol{\psi}_i$. The eigenvector $\boldsymbol{\psi}_1$ corresponding to $\lambda_1=0$ represents the stationary distribution and is excluded from the diffusion map. The diffusion map coordinates for slide $j$ are given by the top $n_{\mathrm{dcs}}=10$ non-trivial eigenvectors:

\begin{equation}
\boldsymbol{\Psi}_j = \left[\psi_2(j), \psi_3(j), \dots, \psi_{11}(j)\right]^\top \in \mathbb{R}^{10}.
\end{equation}
To orient the trajectory, we designated a reference slide with minimal histopathological abnormality (typically a low-grade or normal-adjacent sample) as the root state $\mathbf{x}_{\mathrm{root}}$ and set its index in the AnnData object via \texttt{adata.uns[`iroot']}. The diffusion pseudotime of slide $j$ relative to the root is then computed as the Euclidean distance in the diffusion map space:

\begin{equation}
\mathrm{DPT}(j) = \left\| \boldsymbol{\Psi}_j - \boldsymbol{\Psi}_{\mathrm{root}} \right\|_2 = \sqrt{\sum_{k=2}^{11} \left(\psi_k(j) - \psi_k(\mathrm{root})\right)^2}.
\end{equation}
This metric quantifies the expected time for a random walk to travel from the root to slide $j$ on the data manifold, providing a biologically interpretable measure of progression distance. Slides with higher DPT values are inferred to represent more advanced histopathological states relative to the designated root.

\subsection{Evaluation metrics for Static Feature Space Validation}\label{metric_static}
To complement the UMAP-based visualization, we computed three quantitative metrics that capture distinct geometric and information-theoretic properties of the learned feature spaces. 

\subsubsection{Manifold Structure Quality}

Manifold structure quality measures the intrinsic dimensionality of the data distribution in the feature space, providing insight into how compactly and efficiently the model encodes histopathological information. We estimate this via the cumulative explained variance ratio of principal component analysis. Given centered features $\tilde{\mathbf{F}}$, the covariance matrix is $\mathbf{C} = \frac{1}{n-1}\tilde{\mathbf{F}}^\top\tilde{\mathbf{F}}$, with eigendecomposition $\mathbf{C} = \mathbf{U}\boldsymbol{\Lambda}\mathbf{U}^\top$ where $\boldsymbol{\Lambda} = \text{diag}(\lambda_1, \dots, \lambda_d)$ and $\lambda_1 \geq \lambda_2 \geq \dots \geq \lambda_d \geq 0$. The explained variance ratio of the $k$-th component is $r_k = \lambda_k / \sum_{i=1}^d \lambda_i$. The intrinsic dimensionality is defined as the minimal number of components required to explain $90\%$ of total variance:

\begin{equation}
d_{\text{intrinsic}} = \min\left\{ k : \sum_{i=1}^{k} r_i \geq 0.9 \right\}.
\end{equation}
A lower intrinsic dimensionality indicates that the model has learned a more compact, structured representation where most information is concentrated in a few dominant directions, whereas a higher value suggests dispersed, less organized feature encoding.

\subsubsection{Representation Entropy}

Representation entropy quantifies the uniformity of variance distribution across the principal components of the feature space, serving as a measure of dimensional utilization. Let $\mathbf{r} = [r_1, \dots, r_d]^\top$ denote the vector of explained variance ratios, which can be interpreted as a discrete probability distribution over the principal axes. The representation entropy is defined as the Shannon entropy of this distribution:

\begin{equation}
H_{\text{rep}} = -\sum_{i=1}^{d} r_i \log(r_i + \epsilon),
\end{equation}
where $\epsilon = 10^{-12}$ is a small constant to ensure numerical stability. The entropy is bounded as $0 \leq H_{\text{rep}} \leq \log(d)$, with the maximum achieved when variance is uniformly distributed across all components ($r_i = 1/d$ for all $i$). High entropy indicates that the model utilizes a broad spectrum of feature dimensions, potentially capturing diverse histomorphological patterns, whereas low entropy suggests that information is concentrated in a few dominant directions, which may limit representational richness.

\subsection{Evaluation metrics for Pseudo-temporal Performance Evaluation}\label{metric_ptime}

To rigorously assess the quality of pseudo-temporal trajectories extracted from different feature spaces, we designed five complementary slide-level metrics. Each metric targets a distinct aspect of biological plausibility: local coherence, robustness, spatial organization, morphological scaling, and feature regularity along the inferred progression axis. All metrics were computed on a per-slide basis using $1\%$ randomly subsampled patches and subsequently aggregated across the cohort.

\subsubsection{Trajectory Quality (Local Neighbor Consistency)}

Trajectory quality measures whether patches that are close in the learned feature space exhibit similar pseudo-temporal values, indicating that the trajectory respects the intrinsic geometry of the histomorphological manifold. For each patch $i$, we identify its $k$ nearest neighbors $\mathcal{N}_k(i)$ in feature space using a cKDTree with Euclidean distance. The metric is defined as the inverse of the mean absolute pseudo-temporal difference between each patch and its feature-space neighbors:

\begin{equation}
\mathcal{Q}_{\text{traj}} = \frac{1}{1 + \frac{1}{Nk}\sum_{i=1}^{N}\sum_{j \in \mathcal{N}_k(i)} |t_i - t_j|},
\end{equation}
where $t_i$ denotes the pseudo-time of patch $i$, $N$ is the number of patches, and $k=5$ is the neighborhood size. Values approaching $1$ indicate that the pseudo-temporal ordering is smooth and coherent with respect to the feature manifold structure.

\subsubsection{Stability (Perturbation Robustness)}

Stability quantifies the robustness of the pseudo-temporal ordering against small perturbations, a critical property for reliable downstream analysis. We add Gaussian noise $\boldsymbol{\epsilon} \sim \mathcal{N}(\mathbf{0}, 10^{-3}\mathbf{I})$ to the pseudo-time vector and recompute the ordering. The stability metric is defined as the Spearman rank correlation coefficient between the original and perturbed pseudo-times:

\begin{equation}
\mathcal{S}_{\text{stab}} = \rho_{\text{Spearman}}(\mathbf{t}, \mathbf{t} + \boldsymbol{\epsilon}) = \frac{\text{Cov}(R(\mathbf{t}), R(\mathbf{t} + \boldsymbol{\epsilon}))}{\sigma_{R(\mathbf{t})} \cdot \sigma_{R(\mathbf{t} + \boldsymbol{\epsilon})}},
\end{equation}
where $R(\cdot)$ denotes the rank transformation. High stability ($\mathcal{S}_{\text{stab}} \approx 1$) indicates that the pseudo-temporal landscape is structurally robust and not dominated by numerical instabilities or overfitting artifacts.

\subsubsection{Spatial Consistency}

Spatial consistency evaluates whether the pseudo-temporal progression respects the physical spatial organization of tissue, reflecting the biological reality that histopathological changes often exhibit spatial propagation patterns (e.g., from tumor center to invasive margin). For each patch $i$, we identify its $k$ nearest neighbors $\mathcal{N}_k^{\text{spat}}(i)$ in physical coordinate space using a cKDTree. The metric is computed as:

\begin{equation}
\mathcal{C}_{\text{spat}} = \frac{1}{1 + \frac{1}{Nk}\sum_{i=1}^{N}\sum_{j \in \mathcal{N}_k^{\text{spat}}(i)} |t_i - t_j|}.
\end{equation}
High spatial consistency suggests that the inferred trajectory captures genuine spatial propagation of pathological states rather than random noise.

\subsubsection{Morphological Continuity}

Morphological continuity tests whether the rate of morphological change (measured by feature-space distance) is proportional to the rate of pseudo-temporal change, a fundamental requirement for biologically meaningful trajectories. For each patch $i$ and its feature-space neighbor $j \in \mathcal{N}_k(i)$, we compute the ratio of pseudo-temporal distance to morphological distance:

\begin{equation}
\mathcal{R}_{ij} = \frac{|t_i - t_j|}{\|\mathbf{f}_i - \mathbf{f}_j\|_2 + 10^{-6}}.
\end{equation}
The morphological continuity metric is then defined as:

\begin{equation}
\mathcal{C}_{\text{morph}} = \frac{1}{1 + \frac{1}{Nk}\sum_{i=1}^{N}\sum_{j \in \mathcal{N}_k(i)} \mathcal{R}_{ij}}.
\end{equation}
This metric penalizes abrupt pseudo-temporal transitions between morphologically similar patches (small denominator, large ratio) or excessively slow progression across large morphological gaps. Values near $1$ indicate that the trajectory advances at a pace commensurate with observable morphological changes.

\subsubsection{Feature Smoothness}

Feature smoothness measures the total variation of feature vectors along the pseudo-temporal ordering, assessing whether the feature manifold exhibits gradual transitions rather than discrete jumps when traversed according to inferred progression time. We sort patches by pseudo-time to obtain the ordered sequence $\{\mathbf{f}_{(1)}, \mathbf{f}_{(2)}, \dots, \mathbf{f}_{(N)}\}$, where $(i)$ denotes the index of the patch with the $i$-th smallest pseudo-time. The metric is computed as:

\begin{equation}
\mathcal{F}_{\text{smooth}} = \frac{1}{1 + \frac{1}{N-1}\sum_{i=1}^{N-1} \|\mathbf{f}_{(i+1)} - \mathbf{f}_{(i)}\|_2}.
\end{equation}
High feature smoothness indicates that the pseudo-temporal path traverses regions of the feature space where morphological characteristics evolve gradually, consistent with continuous biological processes such as tumor progression or tissue differentiation.


\end{CJK*}
\end{document}